\documentclass{article}

\usepackage{microtype}
\usepackage{graphicx}
\usepackage{subcaption}
\usepackage{booktabs} % for professional tables

\usepackage{hyperref}

\usepackage[preprint]{icml2026}

\usepackage{amsmath}
\usepackage{amssymb}
\usepackage{mathtools}
\usepackage{amsthm}

\usepackage[capitalize,noabbrev]{cleveref}

\usepackage{multicol}
\usepackage{multirow}
\usepackage[T1]{fontenc}
\usepackage{xcolor}          % 提供彩色
\newcommand{\gaing}[1]{\colorbox{gray!15}{+#1}}  % 浅hui底
\newcommand{\gainb}[1]{\colorbox{blue!15}{+#1}}  % 浅蓝底
\newcommand{\gainexp}[1]{\colorbox{blue!15}{#1}}  % 浅蓝底
\usepackage{tcolorbox} 
\usepackage{fvextra} % enhanced Verbatim
\tcbuselibrary{breakable,skins}
\newtcolorbox{promptbox}[2][]{%
  breakable,
  colback=gray!2,
  colframe=gray!40,
  title=\textbf{#2},
  fonttitle=\small,
  sharp corners,
  boxrule=0.5pt,
  left=6pt,right=6pt,top=6pt,bottom=6pt,
  enhanced,
  #1
}
\usepackage{listings}
\lstdefinestyle{prompt}{
  basicstyle=\ttfamily\small,
  columns=fullflexible,
  breaklines=true,
  frame=single,
  rulecolor=\color{gray!40},
  frameround=ffff,
  xleftmargin=1em,
  xrightmargin=1em,
  aboveskip=0.6em,
  belowskip=0.6em
}
\usepackage{caption}

\lstdefinestyle{promptio}{
  basicstyle=\ttfamily\small,
  columns=fullflexible,
  breaklines=true,
  frame=single,
  rulecolor=\color{gray!40},
  frameround=ffff,
  xleftmargin=1em, xrightmargin=1em,
  aboveskip=0.6em, belowskip=0.6em
}
\lstnewenvironment{PromptIO}[2][]{%
  \lstset{style=promptio,caption={#1},label={#2}}%
}{}
\theoremstyle{plain}

\theoremstyle{definition}

\theoremstyle{remark}

\usepackage[textsize=tiny]{todonotes}

\icmltitlerunning{Improving Data and Reward Design for Scientific Reasoning in Large Language Models}

\begin{document}

\twocolumn[
  \icmltitle{Improving Data and Reward Design for Scientific Reasoning \\in Large Language Models}



  \icmlsetsymbol{equal}{*}

  \begin{icmlauthorlist}
    \icmlauthor{Zijie Chen}{zju,wlu,msra}
    \icmlauthor{Zhenghao Lin}{msra}
    \icmlauthor{Xiao Liu}{msra}
    \icmlauthor{Zhenzhong Lan}{wlu}
    \icmlauthor{Yeyun Gong}{msra}
    \icmlauthor{Peng Cheng}{msra}
  \end{icmlauthorlist}

  \icmlaffiliation{zju}{Zhejiang University}
  \icmlaffiliation{wlu}{Westlake University}
  \icmlaffiliation{msra}{Microsoft Research Asia}

  \icmlcorrespondingauthor{Yeyun Gong}{yegong@microsoft.com}

  \icmlkeywords{Machine Learning, ICML}

  \vskip 0.3in
]



\printAffiliationsAndNotice{}  % no special notice (required even if empty)

\begin{abstract}
  % Robust scientific reasoning in large language models (LLMs) remains challenging due to imperfect post-training data curation, misaligned training curricula, and unreliable rewards for open-ended science questions.
  Solving open-ended science questions remains challenging for large language models, particularly due to inherently unreliable supervision and evaluation. The bottleneck lies in the data construction and reward design for scientific post-training.
  We develop a large-scale, systematic data processing pipeline that transforms heterogeneous open-source science data into Dr. SCI dataset, which comprises of 1M questions across eight STEM subjects, with explicit verifiable/open-ended splits, scalable difficulty annotation, and fine-grained rubrics that operationalize evaluation for open-ended answers.
  Building on this dataset, we propose the Dr.\ SCI post-training pipeline, which redesigns the standard SFT$\rightarrow$RL workflow through three components: (i) Exploration-Expanding SFT, which broadens the model’s reasoning pattern coverage prior to RL; (ii) Dynamic Difficulty Curriculum, which adapts training data to the model’s evolving scientific capability; and (iii) SciRubric-Guided RL, which enables stable reinforcement learning on open-ended scientific questions via rubric-based evaluation with explicit answer correctness.
  % We introduce Dr.SCI dataset, a curated scientific reasoning dataset with 1M challenging question spanning eight STEM subjects, with verifiable/open-ended splits, difficulty annotations inferred at scale, and fine-grained rubrics that operationalize evaluation for open-ended questions. 
  % Building on top of this dataset, we introduce the Dr.SCI post-training pipeline. It redesigns the standard SFT$\rightarrow$RL workflow via i) Exploration Expanding SFT: which selects supervision to broaden the model's reasoning-pattern and raise the exploration ceiling for downstream RL; ii) Dynamic Difficulty Curriculum: that continuously adapts RL sampling to the model's evolving capability frontier; and iii) SciRubric-Guided RL: which leverages Dr.SCI dataset's fine-grained rubrics to deliver stable, interpretable rewards for open-ended scientific questions. 
  Qwen3-4B-Base trained using Dr.SCI pipeline achieves 63.2 on GPQA-diamond and 32.4 on GPQA-general, consistently improves over strong post-trained baselines such as o1-mini and GPT-4o, demonstrating substantial gains in scientific reasoning, especially in open-ended settings.
\end{abstract}

%%%%%%%%%%%%%%%%%%%%%%%%%%%%%%%%%%%%%%%%%%%%%%%%%%%%%%%%%%%%%%%%%%%%%%%%%%%%%%%
%%%%%%%%%%%%%%%%%%%%%%%%%%%%%%%%%%%%%%%%%%%%%%%%%%%%%%%%%%%%%%%%%%%%%%%%%%%%%%%
\section{Introduction}

\begin{figure}[t]
    \centering
    \includegraphics[width=0.98\columnwidth]{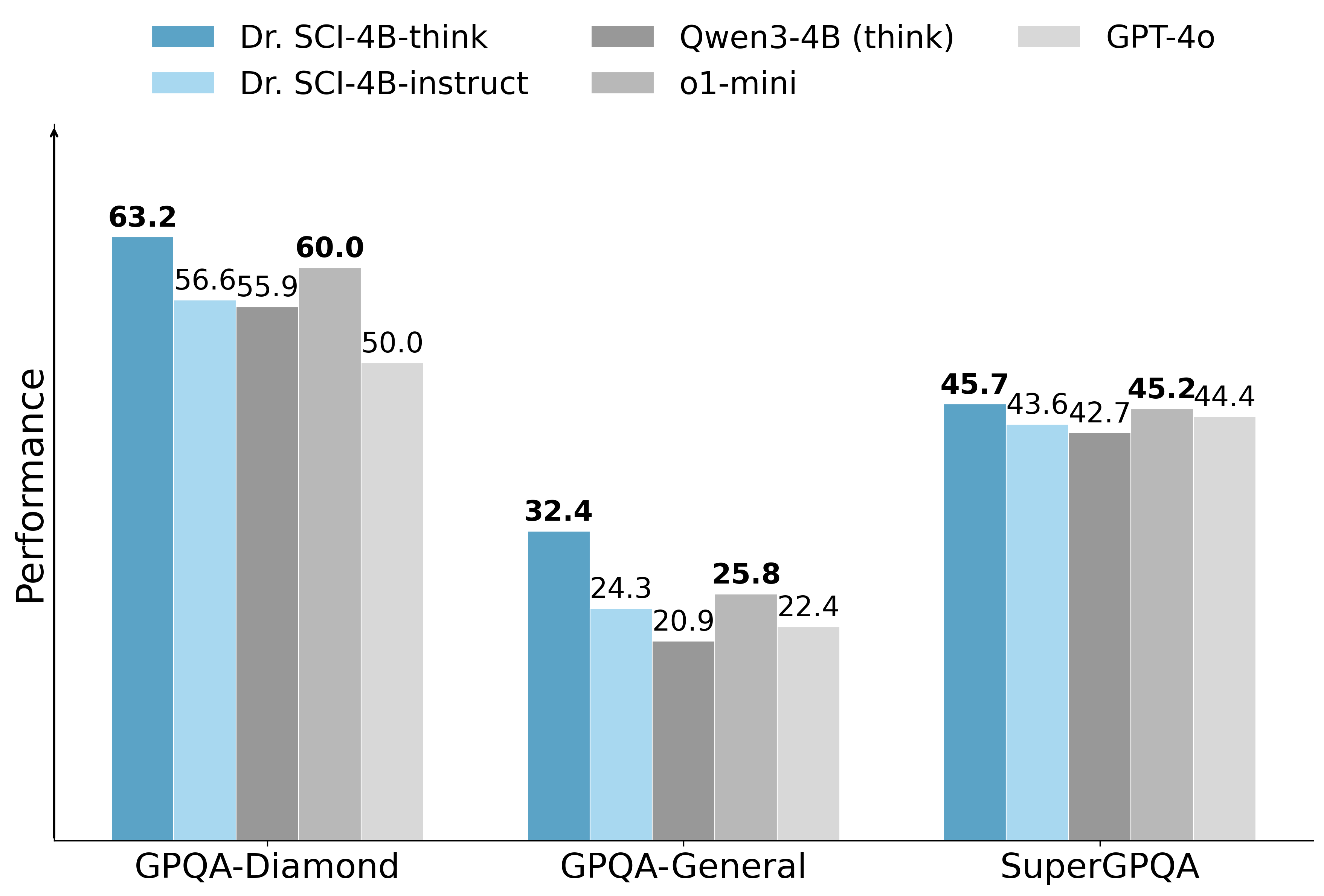}
    \caption{Model performance on core scientific reasoning benchmarks. Dr.\ SCI surpasses strong baselines like o1-mini, GPT-4o.}
    \label{fig: figure 1 performance}
    % \vspace{-3mm}
\end{figure}

Recent advances in large language models (LLMs) have demonstrated strong performance in well-structured reasoning domains such as mathematics~\citep{jaech2024openai,deepseekai2025deepseekr1incentivizingreasoningcapability,hubert2025olympiad}, code~\citep{li2022competition,roziere2023code,luo2023wizardcoder}, and tool- or agent-based tasks~\citep{he2024webvoyager,li2025websailor,5team2025glm45agenticreasoningcoding,team2025kimi} through post training. However, their capabilities remain significantly weaker on open-ended question answering, where answers are often free-form text. Such problems span broad STEM knowledge, involve heterogeneous scientific reasoning patterns and cross-domain generalization~\cite{lu2022learn}. Existing post-training pipelines struggle to reliably elicit high-quality scientific reasoning because supervision and evaluation are inherently unreliable for open-ended science: references are free-form, automatic verification is difficult, and naive rule/string matching fails. These limitations are especially harmful for reinforcement learning (RL), where effective optimization critically depends on stable, informative, and well-defined reward signals.

We argue that a core bottleneck lies in the data construction and evaluation design for scientific post-training. Most open-source science datasets~\cite{fan2025megascience,yuan2025naturalreasoning,guha2025openthoughts} are constructed through loosely controlled pipelines that but vary substantially in their supervision design. Some datasets~\citep{guha2025openthoughts,NemotronPostTrainingDatasetV1} are primarily distilled for supervised fine-tuning (SFT), providing teacher rationales instead of reference answers; while others~\cite{fan2025megascience,yuan2025naturalreasoning} approximate open-ended tasks but without reliable verification guidance.
 % These shortcomings are most severe for open-ended scientific questions, which constitute a large fraction of real-world scientific reasoning tasks but lack a principled mechanism for automated evaluation.
In addition, difficulty annotation is largely absent, leaving many instances too easy or inappropriate for curriculum learning.  As a result, existing datasets are poorly aligned with the needs of scientific post-training.

To address this gap, we develop a  principled data processing pipeline for scientific reasoning, which transforms heterogeneous open-source resources into the Dr.\ SCI dataset. Dr.\ SCI comprises 1,006,701 curated problems across eight STEM subjects, with explicit partitioning into 461K rule-verifiable and 545K open-ended instances, rigorous quality control, and scalable difficulty annotation to support reliable post-training and adaptive curricula. Dr.\ SCI also provides structured supervision for open-ended scientific questions via carefully generated evaluation rubrics. We generate a set of decomposed criteria that characterize a high-quality response for each open-ended instance, serving as a foundation for stable rubric-guided reinforcement learning.

% In this work, we introduce \textbf{Dr.\ SCI}, a unified dataset and post-training pipeline designed explicitly for scientific reasoning. We first construct the \textbf{Dr.\ SCI dataset}, comprising 1,006,701 curated problems across eight STEM subjects, explicitly partitioned into 461K rule-verifiable and 545K open-ended instances. Each question undergoes systematic quality control, scalable difficulty annotation and filtering, and fine-grained evaluation rubrics (14.5 items per question on average) for open-ended questions.

Building on this dataset, we propose the Dr.\ SCI post-training pipeline, which re-engineers each stage to suite scientific reasoning post-training. Exploration-Expanding SFT selects supervision to broaden the model’s reasoning-pattern repertoire prior to RL; a Dynamic Difficulty Curriculum continuously adapts the RL training data to the model’s evolving capability frontier; and SciRubric-Guided RL enables stable optimization on open-ended scientific questions through decomposed rubric-based rewards while explicitly enforcing final-answer correctness. Across diverse scientific reasoning benchmarks, applying Dr.\ SCI to a compact 4B backbone yields substantial gains and consistently outperforms a wide range of strong post-trained baselines including o1-mini\cite{jaech2024openai} and GPT-4o~\cite{hurst2024gpt} as shown in Figure~\ref{fig: figure 1 performance}. Our full data and processing pipeline, training code, and models will be publicly available soon.

We summarize our main contributions as:

1. We develop a large-scale, systematic data processing pipeline that transforms heterogeneous open-source science data into \textbf{Dr.\ SCI} dataset, featuring explicit rule-verifiable and open-ended splits, rigorous quality control, and scalable difficulty annotation to support reliable post-training.

2. We generate fine-grained rubrics for open-ended scientific questions in Dr.\ SCI dataset, and propose \textbf{SciRubric-Guided RL}, which leverages these rubrics together with explicit final-answer correctness to provide stable and informative reward signals for reinforcement learning.

3. We integrate \textbf{Exploration-Expanding SFT}, a \textbf{Dynamic Difficulty Curriculum}, and \textbf{SciRubric-Guided RL} into a coherent post-training pipeline, systematically improving exploration and optimization dynamics and achieving substantial scientific reasoning gains from a compact 4B backbone, surpassing much larger post-trained models.

%%%%%%%%%%%%%%%%%%%%%%%%%%%%%%%%%%%%%%%%%%%%%%%%%%%%%%%%%%%%%%%%%%%%%%%%%%%%%%%
%%%%%%%%%%%%%%%%%%%%%%%%%%%%%%%%%%%%%%%%%%%%%%%%%%%%%%%%%%%%%%%%%%%%%%%%%%%%%%%
\section{Dr. SCI Dataset}
\label{sec: dataset}
We introduce Dr.\ SCI dataset, a large-scale scientific reasoning dataset collected using a systematic data processing pipeline. Our pipeline transforms heterogeneous open-source scientific corpora into a well-structured dataset comprising 1,006,701 questions across eight STEM subjects, and is augmented with verification structure, scalable difficulty annotation, and fine-grained rubrics for open-ended questions to aid support scientific post-traing and evaluation.

% We introduce Dr.SCI dataset, a large-scale dataset of scientific reasoning questions designed for training and evaluating large language models at scale. It comprises 1 million carefully curated problems spanning mathematics, physics, chemistry, biology, medicine, computer science, and economics, with systematic quality control, rich metadata annotations and fine-grained rubrics for open-ended questions to aid verification.

\subsection{Data Collection}

We start from high-quality, publicly available scientific datasets, including WebInstruct-Verified~\cite{ma2025general}, NaturalReasoning~\cite{yuan2025naturalreasoning}, MegaScience~\cite{fan2025megascience}, and RaR-Science~\cite{gunjal2025rubrics}. These sources cover a wide range of STEM domains and problem formats, drawing from textbooks, scientific literature, and authoritative website resources.

However, existing science datasets are often built with disparate goals and insufficient preprocessing, resulting in unclear domain partitioning and inconsistent supervision design. Most resources focus primarily on verifiable questions, while open-ended scientific problems lack reliable evaluation structure and are thus unsuitable for direct use in reinforcement learning. These limitations make raw aggregated data misaligned with the requirements of modern scientific post-training.

\subsection{Data Processing Pipeline}

\begin{figure*}[h]
    \centering
    \includegraphics[width=0.9\textwidth]{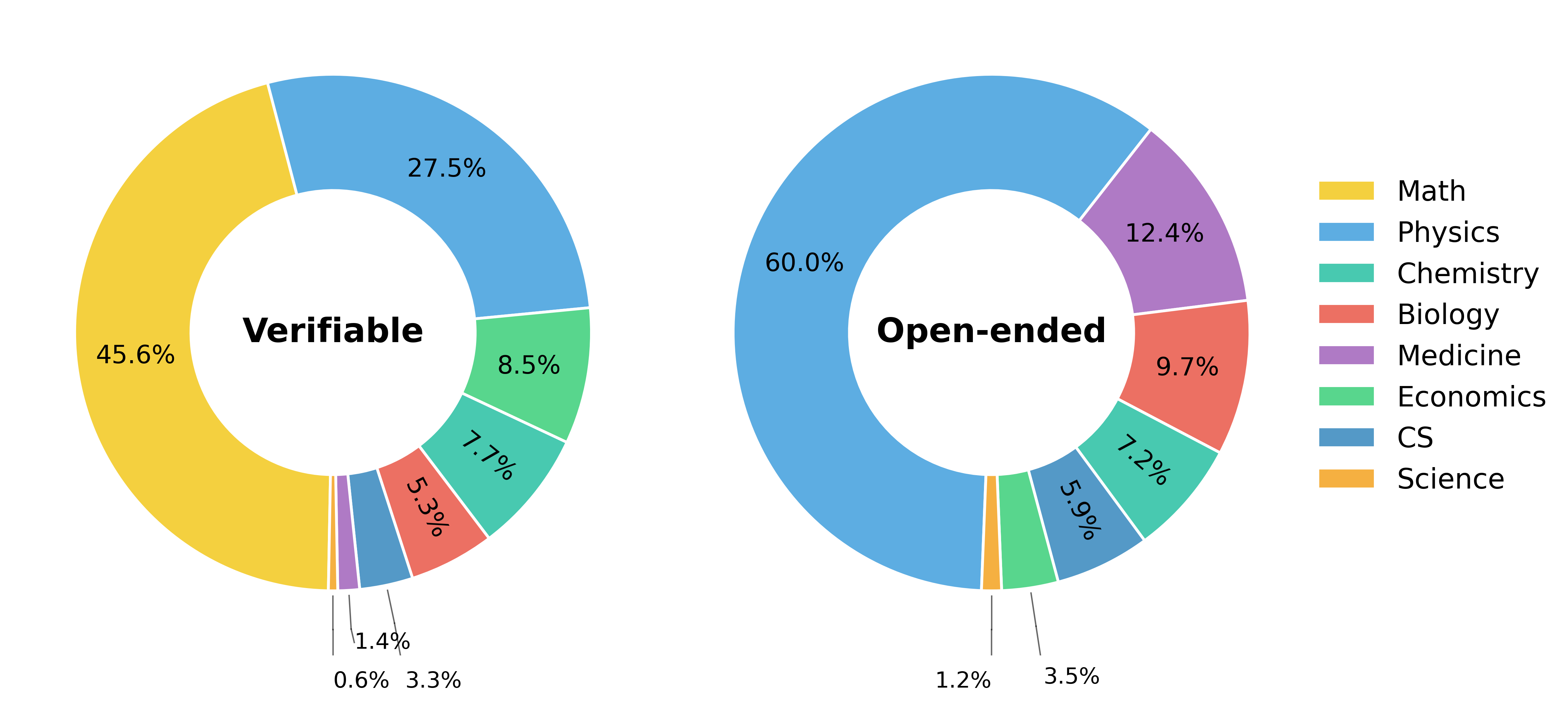}
    \caption{Subject distribution of Dr.\ SCI dataset.}
    \label{fig:dataset subject}
    % \vspace{-3mm}
\end{figure*}

To address these issues, we develop a scalable data processing pipeline that systematically cleans, structures, and augments the collected data. We first remove samples with empty or malformed reference answers, and assign each remaining question to one of seven STEM subjects: mathematics, physics, chemistry, biology, medicine, computer science, and economics. Questions that are clearly STEM-related but do not fit cleanly into any of these categories are labeled as the general \textit{science} domain.

We then partition questions into two mutually exclusive classes: \textit{verifiable} and \textit{open-ended}. A question is considered verifiable if its reference answer admits deterministic validation (e.g., numerical values, mathematical expressions, or multiple-choice keys); all others are categorized as open-ended. For verifiable questions, reference answers are canonicalized into minimal checkable forms. We discard open-ended questions in mathematics, as they are predominantly proof-based and empirically induce overlong responses during post-training.

Then, the dataset is deduplicated via exact and near-duplicate matching. Conflicting instances with identical questions but inconsistent reference answers are resolved through answer-equivalence verification, and contaminated samples overlapping with evaluation benchmarks in Section~\ref{sec: experiment evaluation} are removed to ensure reliable generalization.

Next, we estimate question difficulty using the non-thinking version of Qwen3-32B~\cite{yang2025qwen3}. For each question, we perform eight independent rollouts and use the success rate as a difficulty proxy. Verifiable questions are evaluated via rule-based checkers, while open-ended questions are assessed using a generative verifier with prompts specified in Appendix~\ref{appn: prompt final answer}. 413K Questions solved in all attempts (8/8) are discarded as trivial, yielding the final Dr.\ SCI dataset of 1,006,701 instances.

To support structured supervision for open-ended scientific reasoning, we further generate fine-grained evaluation rubrics for all open-ended questions. We prompt OpenAI o3~\cite{o3} to analyze each question, and attempt a solution when necessary, to identify the key criteria that characterize a high-quality response (see Appendix~\ref{appn: prompt rubric gen}). Each question is paired with 7–20 atomic rubric items, each labeled by importance as:

(i) \textit{Essential}: critical fact or step; omission invalidates the answer.

(ii) \textit{Important}: key information or reasoning; absence severely weakens the response.

(iii) \textit{Optional}: secondary details or actions; doesn't directly affects correctness.

(iv) \textit{Pitfall}: common vital mistakes that must be penalized.

Overall, this produces an average of 14.5 rubric items per open-ended question, including 4.3 Essential items, forming the basis for rubric-guided reinforcement learning. An example open-ended question and its corresponding rubrics are provided in the Appendix~\ref{appn: example drsci}.

\subsection{Dataset Statistics}

Dr.\ SCI dataset contains 1,006,701 challenging scientific reasoning questions spanning eight STEM subjects, with subject distributions shown in Figure~\ref{fig:dataset subject}. Although mathematics and physics dominate many real-world sources, Dr.\ SCI maintains broad coverage across domains: each of the remaining subjects contributes more than 47K instances, ensuring diverse scientific concepts and problem formats.

% Dr.\ SCI dataset explicitly supports both \textit{rule-verifiable} and \textit{open-ended} supervision. The dataset comprises 461K verifiable and 545K open-ended questions within a unified resource, enabling complementary training  regimes including rule-based verification and rubric-guided assessment for open-ended reasoning.
Dr.\ SCI dataset explicitly supports both \emph{rule-verifiable} and \emph{open-ended} supervision, with 461K verifiable and 545K open-ended questions respectively. This enables rule-based and rubric-guided assessment in a unified training regime.

\begin{figure}[t]
    \centering
    \includegraphics[width=0.95\columnwidth]{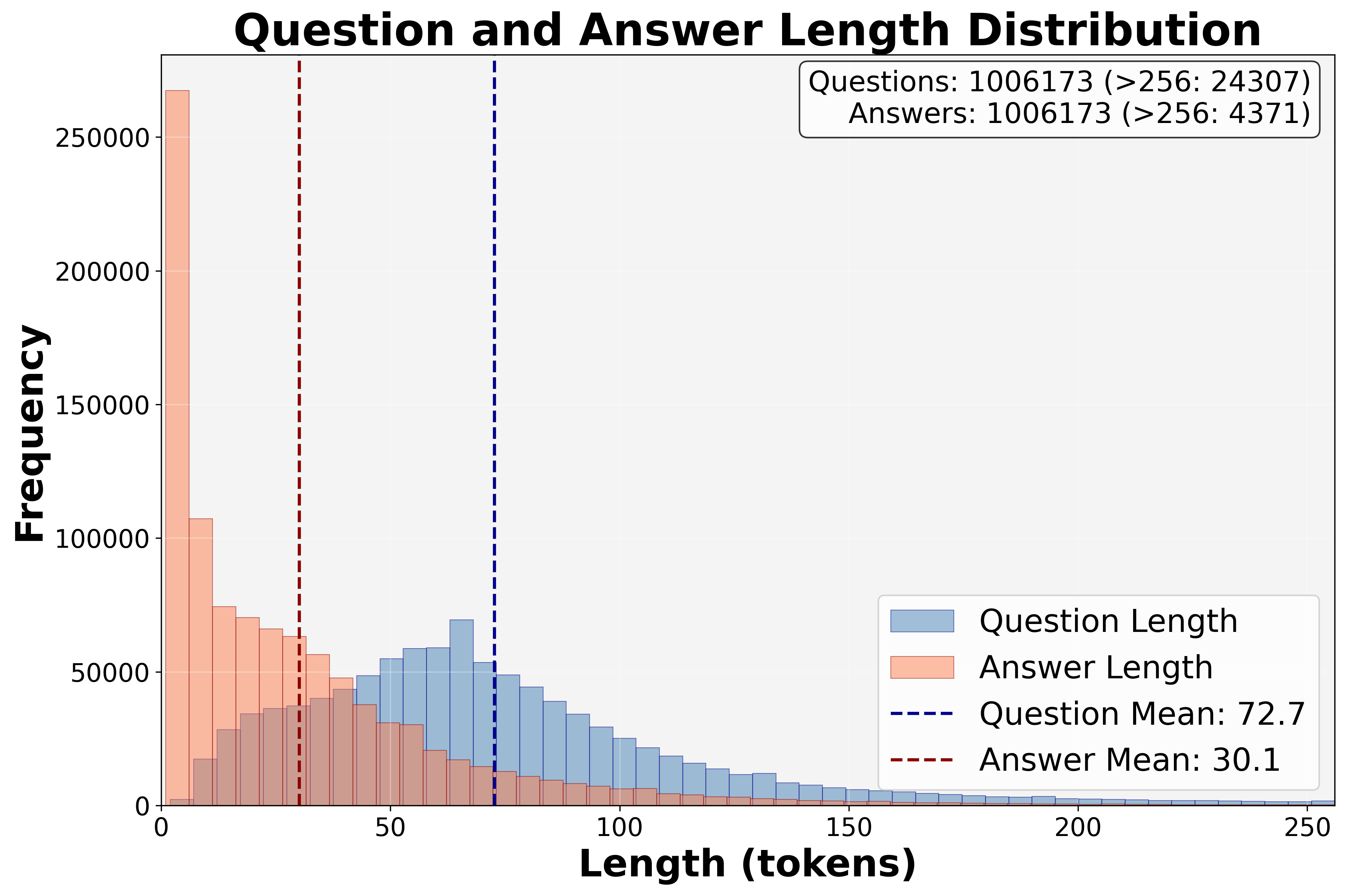}
    \caption{Length Distribution of Dr.\ SCI dataset.}
    \label{fig:dataset length}
    % \vspace{-5mm}
\end{figure}

Figure~\ref{fig:dataset length} summarizes question and answer length distributions. Questions average 72.7 tokens, and reference answers  average 30.1 tokens, facilitating efficient automated verification and large-scale RL training. Only 0.3\% of reference answers exceed 250 tokens, primarily reflecting complex open-ended explanations, preserving necessary difficulty without sacrificing overall training efficiency.

\begin{figure}[t]
    \centering
    \includegraphics[width=0.95\columnwidth]{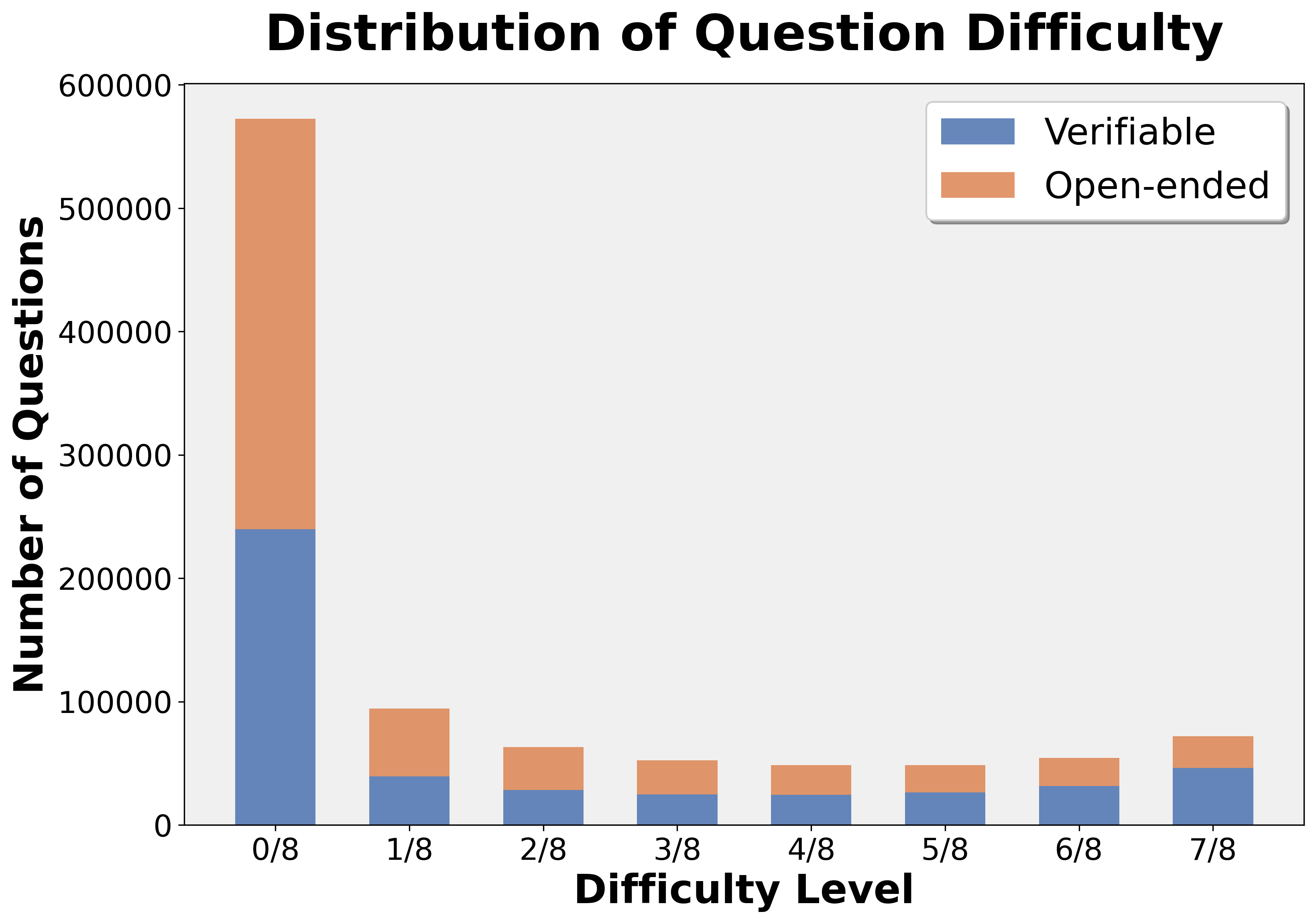}
    \caption{Difficulty Distribution of Dr.\ SCI dataset.}
    \label{fig:dataset difficulty}
    % \vspace{-5mm}
\end{figure}

Finally, Dr.\ SCI exhibits a characteristic J-shaped difficulty distribution (Figure~\ref{fig:dataset difficulty}), consistent with prior observations~\cite{Polaris2025}. The dataset includes abundant hard instances to stress scientific reasoning, while retaining easier examples that stabilize early learning and support curriculum-based training.

Taken together, Dr.\ SCI couples broad STEM coverage, explicit verifiable/open-ended splits, compact lengths, and a calibrated difficulty profile, serving a reliable foundation for large-scale scientific post-training.

% Taken together, these statistics highlight several properties that make Dr.\ SCI well-suited for scientific post-training: broad STEM coverage with a large fraction of open-ended questions, filtered difficulty to suite model LLM post-training. These characteristics, together with explicit verifiable/open-ended partitioning, distinguish Dr.\ SCI from prior aggregated science datasets and provide a reliable foundation for large-scale reinforcement learning.

\section{Dr. SCI Post Training}

Existing large-scale reasoning post-training pipelines typically follow a two-stage recipe: supervised fine-tuning (SFT) on teacher-generated responses, followed by reinforcement learning with verifiable rewards (RLVR). 
While effective for structured domains, this paradigm is poorly suited for scientific reasoning, which is dominated by open-ended questions whose solutions are expressed in free-form explanations and lack reliable verification signals. As a result, both supervision and reinforcement learning become difficult to apply in a stable and principled manner.
% While effective for structured domains, this paradigm is poorly suited for scientific reasoning. First, SFT data selection guided by heuristic filters (e.g., reject sampling or voting) fail to explicitly promote diverse reasoning behaviors. Second, RL curricula are commonly static and poorly aligned with the model’s rapidly evolving capability. Third and most critically, the prevalence of open-ended scientific questions makes RL difficult to apply due to unstable or under-specified reward signals.

We propose \textbf{Dr.\ SCI}, a holistic post-training pipeline that redesigns each stage to address these challenges and explicitly optimize downstream RL performance. Our approach integrates three complementary components: 
(\emph{i})~\textbf{Exploration-Expanding SFT}, which selects supervision to broaden the model’s reasoning-pattern repertoire prior to RL; 
(\emph{ii})~\textbf{Dynamic Difficulty Curriculum}, which continuously adapts the training distribution to the model’s current capability frontier; and 
(\emph{iii})~\textbf{SciRubric-Guided RL}, which enables stable reinforcement learning on open-ended scientific questions through fine-grained, criterion-based evaluation with explicit final-answer correctness. 
Together, these components form a unified post-training pipeline that scales reinforcement learning to open-ended scientific reasoning.

% Conventional large-scale reasoning post-training typically applies supervised fine-tuning (SFT) on teacher-generated responses, followed by reinforcement learning with verifiable rewards (RLVR). However, this recipe falls short for scientific reasoning: SFT data selection is often driven by heuristic filters (e.g., reject sampling or voting) that do not explicitly promote diverse reasoning behaviors; RL curricula are commonly static or poorly aligned with the model’s rapidly evolving capability; and, crucially, the abundant \textit{open-ended} scientific questions remain difficult to optimize with RL due to unstable or under-specified reward signals.

% We present Dr.\ SCI, a holistic post-training pipeline that re-engineers each stage to explicitly maximize downstream RL performance across the full spectrum of scientific reasoning. It integrates three synergistic components: 
% (\emph{i})~\textbf{Exploration-Expanding SFT}, which encourages broad exploration of reasoning patterns before RL based on a quantitative assessment of diversity; (\emph{ii})~\textbf{Dynamic Difficulty Curriculum}, which continuously adjusts the training distribution to match the model’s current capability frontier during RL; and (\emph{iii})~\textbf{SciRubric-Guided RL}, which enables stable optimization on open-ended questions via fine-grained, decomposed rubric-based evaluation. 
% Together, these components unlock scalable RL for scientific reasoning, bridging the gap between deterministic, rule-verifiable supervision and the open-ended complexity of real-world scientific problem solving.

\subsection{Exploration-Expanding SFT}
\label{sec: method 4gram}
% Effective RL relies on sufficient exploration: the policy must generate diverse reasoning behaviors for RL to discover and reinforce high-reward trajectories. 
Scientific questions often require diverse reasoning strategies and explanation styles across domains~\cite{lu2022learn}, making exploration particularly critical. To avoid constraining downstream RL to a narrow reasoning regime, we deliberately broaden the model’s reasoning repertoire during SFT to raise the exploration ceiling for scientific RL.
% When SFT supervision covers only a narrow set of reasoning patterns, subsequent RL is constrained to a limited behavior space, reducing its effectiveness. We therefore aim to broaden the model’s reasoning repertoire during SFT to raise RL ceiling.

To assess reasoning-pattern diversity, we adopt a simple lexical proxy based on 4-gram novelty. 
For each candidate example $d$, let $g(d)$ be its 4-gram set. Given a selected SFT set $\mathcal{D}^*$, we define $\mathcal{G}^*=\bigcup_{d\in\mathcal{D}^*} g(d)$ and use the number of unique 4-grams $|\mathcal{G}^*|$ as the coverage measure. This provides a scalable and model-agnostic signal that correlates with exposure to diverse reasoning traces.
% For each candidate training example $d$, we extract its set of 4-grams $g(d)$. Given a selected SFT dataset $\mathcal{D}^*$, we define the cumulative 4-gram set $\mathcal{G}^*=\bigcup_{d\in\mathcal{D}^*} g(d)$ and use $|\mathcal{G}^*|$ as a measure of local pattern coverage. While coarse, this proxy provides a scalable and model-agnostic signal that correlates with exposure to diverse reasoning traces.

We construct candidate SFT data from a pool of questions $\mathcal{Q}$ drawn from Dr.\ SCI, specifically MegaScience~\cite{fan2025megascience} and WebInstruct-Verified~\cite{ma2025general}. For each question, we generate multiple candidate responses using a diverse set of open source models, covering both thinking (e.g. DeepSeek-R1-0528~\cite{deepseekai2025deepseekr1incentivizingreasoningcapability}) and instruct (e.g. GLM-4.6~\cite{5team2025glm45agenticreasoningcoding}) modes.

\begin{algorithm}[t]
\caption{Exploration-Expanding SFT}
\label{alg: 4gram_selection}
\begin{algorithmic}
\STATE {\bfseries Input:} {Dataset pool $\mathcal{D}$, target size $N$, base model $\pi_{base}$}
\STATE {\bfseries Output:} {Selected dataset $\mathcal{D}^*$, fine-tuned model $\pi_{0}$}
\STATE $\mathcal{D}^* \leftarrow \emptyset$
\STATE $\mathcal{G}^* \leftarrow \emptyset$ \hfill \(\triangleright\)~Cumulative selected 4-grams 
\FOR{$n = 1$ {\bfseries to} $N$}
    \STATE $d^* \leftarrow \arg\max_{d \in \mathcal{D}\setminus \mathcal{D}^*} |g(d) \setminus \mathcal{G}^*|$
    \STATE $\mathcal{D}^* \leftarrow \mathcal{D}^* \cup \{d^*\}$
    \STATE $\mathcal{G}^* \leftarrow \mathcal{G}^* \cup g(d^*)$
\ENDFOR
\STATE $\pi_0 \leftarrow \textrm{SFT}(\mathcal{D}^*,\pi_{base})$
\STATE {\bfseries return}  $\pi_0, \mathcal{D}^{*}$
\end{algorithmic}
\end{algorithm}

Given a target size $N$, we select a subset $\mathcal{D}^*\subseteq\mathcal{D}$ by greedily maximizing incremental 4-gram coverage (Algorithm~\ref{alg: 4gram_selection}). At each step, we choose the example that contributes the largest number of previously unseen 4-grams relative to the current $\mathcal{G}^*$. This procedure favors examples that expand pattern coverage, producing an SFT dataset with higher lexical and structural diversity.

Finally, we fine-tune our base model on the selected subsets $\mathcal{D}^*_{\text{think}}$ and $\mathcal{D}^*_{\text{inst}}$, yielding two initial policies for subsequent RL. By explicitly expanding reasoning-pattern coverage during SFT, we improve exploration and enables more effective reinforcement learning in later stages for base model.

% Given a target size $N$, we select a subset $\mathcal{D}^*\subseteq\mathcal{D}$ by greedily maximizing the number of \emph{new} 4-grams introduced at each step (Algorithm~\ref{alg: 4gram_selection}). Specifically, at iteration $n$ we choose the example whose 4-gram set contributes the largest number of previously unseen 4-grams relative to the current cumulative set $\mathcal{G}^*$. We repeat until $|\mathcal{D}^*|=N$. This greedy procedure ensures that each selected example contributes maximal incremental novelty, expanding the overall pattern repertoire of the SFT dataset.

% Finally, we fine-tune the base model on the selected subsets $\mathcal{D}^*_{\text{think}}$ and $\mathcal{D}^*_{\text{inst}}$, producing two initial policies for the subsequent RL stage (thinking mode and instruct mode, respectively). By explicitly encouraging diverse local patterns in SFT, Exploration-Expanding SFT increases the behavioral diversity available to RL, enabling more effective exploration and recombination of reasoning patterns during optimization.

\subsection{Dynamic Difficulty Curriculum}
\label{sec: method curriculum}

Scientific reasoning datasets are inherently imbalanced, with many simple factual or near-trivial questions coexisting with substantially harder problems that demand complex scientific reasoning. Repeatedly training on easy instances yields diminishing returns, while naive exposure to difficult questions leads to weak and unstable learning signals. We therefore introduce a dynamic curriculum that continuously adapts the training distribution to the model’s current scientific reasoning capability.

% Effective reinforcement learning requires non-degenerate reward signals~\citep{yu2025dapo,Polaris2025}. When training data is overly easy, rewards quickly saturate; when it is too hard, rewards collapse toward zero. In both cases, rewards and advantages become poorly differentiated across samples, yielding weak learning signals and substantially reducing the effectiveness of RL. To avoid these extremes, we introduce a dynamic curriculum that adapts in real time to the model’s evolving capability.

% Effective RL relies on non-degenerate reward signals. When data is too easy or too hard, rewards become saturated or collapsed, weakening learning progress.
% To avoid wasting compute at either extreme, we introduce a curriculum that \emph{adapts in real time} to the model’s evolving capability.

% Assume each training sample $x\in\mathcal{D}$ is annotated with a difficulty proxy $d(x)\in\{0,1,\ldots,8\}/8$ as in our dataset construction: larger $d(x)$ indicates the sample is \emph{easier} (e.g., $8/8$ means consistently solved), while smaller $d(x)$ indicates a \emph{harder} sample (e.g., $0/8$ means consistently failed). We partition $\mathcal{D}$ into three subsets:
Each training sample $x \in \mathcal{D}$ is associated with a difficulty $d(x) \in \{0,1,\ldots,8\}/8$, estimated during construction of Dr.\ SCI dataset, where larger values indicate easier instances. Using this signal, we partition $\mathcal{D}$ into three subsets:

(i) $\mathcal{D}_{\text{discard}} = \{x \mid d(x) \ge \tau_{\text{discard}}\}$, consisting of trivial instances removed from RL training.

(ii) $\mathcal{D}_{\text{pending}} = \{x \mid d(x) \le \tau_{\text{pending}}\}$, consisting of currently too-difficult instances deferred for later training.

(iii) $\mathcal{D}_{\text{train}} = \mathcal{D} \setminus (\mathcal{D}_{\text{discard}} \cup \mathcal{D}_{\text{pending}})$, which forms the inital active training set.

We set $\tau_{\text{discard}} = 1.0$ and $\tau_{\text{pending}} = 0.625$ by default.

RL is initialized on $\mathcal{D}_{\text{train}}$. During training, we track an average rollout accuracy $acc(x)$ for each question within the current epoch. If $acc(x)$ is larger than a threshold $\tau_{\text{train}}$, the sample is considered \emph{mastered} and marked for replacement. We set $\tau_{\text{train}}=0.9$ in practice. At the end of each epoch, each mastered sample is replaced with an instance drawn from the easiest remaining subset of $\mathcal{D}_{\text{pending}}$, i.e.,

% We initialize RL on $\mathcal{D}_{\text{train}}$ and track an empirical rollout accuracy $acc(x)$ for each sample within the current epoch. If a sample achieves $acc(x) > \tau_{\text{train}}$ (default $\tau_{\text{train}}=0.9$), we treat it as \emph{mastered} and mark it for replacement. At the end of each epoch, we replace each marked instance with a randomly chosen sample from the \emph{easiest stratum} of $\mathcal{D}_{\text{pending}}$, i.e.,

% \vspace{-3mm}
\begin{center}
$x \sim \left\{x' \in \mathcal{D}_{\text{pending}} \;\middle|\; d(x') = \min_{z\in\mathcal{D}_{\text{pending}}} d(z)\right\}$
\end{center}
% \vspace{-3mm}

This curricula gradually increases training difficulty as the model improves, ensuring that rewards remain informative while avoiding prolonged training on samples that are too easy or hard currently. As a result, the training distribution continuously tracks the model’s capability frontier, improving both data efficiency and final RL performance.

% Intuitively, as the model improves, previously too-hard examples become learnable; by always introducing the easiest remaining instances from $\mathcal{D}_{\text{pending}}$, the curriculum increases difficulty gradually while keeping rewards informative.

% This dynamic replacement strategy maintains a training distribution that continuously matches the model’s capability frontier, improving both data efficiency and final RL performance by avoiding prolonged training on saturated-easy or impossible-hard samples.

% We initialize RL with $\mathcal{D}_{train}$ and continuously track average rollout accuracy $acc(x)$ for each data within current epoch. Examples with $acc(x) > \tau_{train}$ (default $0.9$) indicates already mastered by the model, is marked for replacement. At the end of each epoch, we substitute each marked examples with a random easiest instances from $\mathcal{D}_{\text{pending}}$ i.e., examples with $d(x) = \min_{x'\in\mathcal{D}_{pending}}(d(x'))$. This dynamic curriculum ensures training data continuously matches the model's evolving capability, maximizing RL performance as well as data efficiency.

\subsection{SciRubric-Guided RL}
\label{sec: method rubric}
Open-ended scientific questions pose a fundamental challenge for RL: correctness is rarely determined by simple rules, and naive reference matching yields unstable or uninformative rewards. To address this issue, we leverage the fine-grained rubrics in Dr.\ SCI to provide structured and reliable reward signals for open-ended supervision.

For each open-ended question $x$, Dr.\ SCI provides a reference answer $y_0$ and a set of rubric items $\{r_i\}_{i=1}^{m}$. During RL, for each rollout $y$ generated by the current policy, we extract the final response segment $y_{\text{res}}$ (i.e., the content following ``\texttt{</think>}"). We then evaluate $y_{\text{res}}$ against each rubric item $r_i$ using a lightweight verifier model, producing binary satisfaction indicators $j_i \in \{0,1\}$ that capture whether the response meets each specified criterion.

In addition to rubric-level feedback, we explicitly enforce final-answer correctness. We extract final answer $y_{\text{ans}}$ from $y_{\text{res}}$ by parsing ``\texttt{\textbackslash boxed\{\}}" spans, and compare it with $y_0$ using the same verifier, yielding a binary indicator $j_{\text{ans}} \in \{0,1\}$. This separation ensures that partial rubric satisfaction cannot compensate for an incorrect final answer.

% For each open-ended question $x$ in Dr.\ SCI, we have a reference answer $y_0$ and a set of rubric items $\{r_i\}_{i=1}^{m}$.
% During RL, for each rollout $y$ generated by current policy, we isolate the final response segment $y_{\text{res}}$ (i.e., the content after \texttt{</think>}). We then evaluate $y_{\text{res}}$ against each rubric item $r_i$ using a lightweight verifier model, yielding binary satisfaction indicators $j_i\in\{0,1\}$.

% In addition to rubric-level evaluation, we extract a concise final answer $y_{\text{ans}}$ from $y_{\text{res}}$ via pattern matching of \texttt{\textbackslash boxed\{\}} spans, and assess its correctness against $y_0$ using the same verifier, producing $j_{\text{ans}}\in\{0,1\}$. We define the reward for rollout $y$ as a normalized weighted average of the rubric and answer signals:
We combine rubric satisfaction and final-answer correctness into a single reward signal via a weighted aggregation:
\[
R(y) = \frac{w_{\text{ans}} \cdot j_{\text{ans}} + \sum_{i=1}^{m} w_i \cdot j_i}{w_{\text{ans}} + \sum_{i=1}^{m} w_i},
\]
where $w_{\text{ans}}$ and $w_i$ denote importance weights for final-answer correctness and individual rubric items, respectively, derived from their \textit{Essential}, \textit{Important}, \textit{Optional}, or \textit{Pitfall} categories. Full prompts for  final-answer checking and rubric verification are provided in the Appendix ~\ref{appn: prompt final answer} and~\ref{appn: prompt rubric eval}

This rubric-guided reward provides fine-grained, actionable feedback while maintaining a strong correctness constraint, yielding stable and well-differentiated rewards for open-ended scientific reasoning. When combined with exact, rule-based rewards for verifiable questions, SciRubric-Guided RL enables a unified post-training framework that supports reliable RL across scientific tasks. Empirically, it produces substantially more stable training and stronger performance than prior reward formulations (Section~\ref{sec: ablation Open-Ended}).

% Our rubric-guided reward provides a stable and actionable learning signal for RL on open-ended questions, where correctness is rarely decidable by simple rule-based checkers. Combined with the exact rewards available for \textit{verifiable} instances, this yields a unified post-training framework that leverages both precise automatic verification and fine-grained, decomposed evaluation. Empirically, SciRubric-Guided RL produces stable rewards and substantial gains over baseline reward designs (Section~\ref{sec: ablation Open-Ended}), underscoring its necessity for mastering complex scientific reasoning.

%%%%%%%%%%%%%%%%%%%%%%%%%%%%%%%%%%%%%%%%%%%%%%%%%%%%%%%%%%%%%%%%%%%%%%%%%%%%%%%
%%%%%%%%%%%%%%%%%%%%%%%%%%%%%%%%%%%%%%%%%%%%%%%%%%%%%%%%%%%%%%%%%%%%%%%%%%%%%%%
\section{Experiments}

\subsection{Implementation Details}
\label{sec: experiments implementation}
We adopt Qwen3-4B-Base~\cite{yang2025qwen3} as the base model for scientific reasoning post-training, producing Dr.\ SCI-4B-\emph{think} and Dr.\ SCI-4B-\emph{instruct}.
Unless otherwise specified, we use 1M examples for supervised fine-tuning (SFT) and train for 4 epochs until convergence. Further SFT details in Table~\ref{tab: appn sft hyperparam} in Appendix~\ref{sec: appn implementation detail}.

RL is conducted using the verl~\cite{sheng2024hybridflow} framework with GRPO~\cite{shao2024deepseekmath}. RL runs for up to 10 epochs with the dynamic difficulty curriculum setting as in Section~\ref{sec: method curriculum}. For open-ended questions, we employ Qwen3-4B~\cite{yang2025qwen3} (non-thinking mode) as the verifier with a maximum verification length of 2048 tokens. Rubric item weights are derived from their categories, with final-answer correctness assigned a dominant weight. Full details for RL are reported in Table~\ref{tab: appn RL hyperparam} in Appendix~\ref{sec: appn implementation detail}.

% We conduct RL training using the \texttt{verl} framewor, adopting GRPO~\cite{shao2024deepseekmath} as our algorithm. We train on the SciReason dataset for up to 10 epochs. Difficulty curriculum hyperparameters are set to $\tau_{discard}=1.0$, $\tau_{pending}=0.625$ and $\tau_{train}=0.9$. We use Qwen3-4B non-thinking mode as RM for open-ended questions with the maximum length of each verification as 1024 tokens. The weights for different type of rubrics are: 1.0 for "Essential", 0.7 for "Important", 0.3 for "Optional", 0.9 for "Pitfall" and 4.0 for "Answer Check". Additional RL hyperparameters are provided in Table~\ref{tab: appn RL hyperparam} in Appendix~\ref{appn}.

\subsection{Evaluations}
\label{sec: experiment evaluation}
We evaluate our post-trained models on comprehensive scientific reasoning benchmarks: GPQA-diamond~\cite{rein2024gpqa}, SuperGPQA~\cite{du2025supergpqa}, MMLU-Pro~\cite{wang2024mmlu}, HLE~\cite{phan2025humanity}. To specifically assess open-ended scientific reasoning, we additionally introduce \textbf{GPQA-general}, an open-ended benchmark constructed from GPQA-diamond (See Appendix~\ref{appn: gpqa general construct} for details).

GPQA-general converts all multiple-choice questions in GPQA-diamond into open-ended format by removing answer options and rewriting each question into an unconstrained form using GPT4o~\cite{hurst2024gpt}. As a result, GPQA-general provides the only evaluation in our benchmark suite that measures free-form, open-ended scientific reasoning without reliance on predefined options. 

We report pass@1 for SuperGPQA, HLE, and MMLU-Pro, and avg@10 for GPQA-diamond and GPQA-general. All evaluations use a 32k token context and follow Qwen3 sampling best practices: temp$=0.7$, top-$p=0.8$, top-$K=20$ for instruct models; and temp$=0.6$, top-$p=0.95$, top-$K=20$ for thinking models. For baselines, we evaluate their models under identical settings and report the better result between our runs and those reported in prior work.

% We report pass@1 for SuperGPQA, HLE and MMLU-Pro; avg@10 for GPQA-diamond and GPQA-general. All evaluations use a 32k token context limit, following Qwen3 best practices\footnote{https://huggingface.co/Qwen/Qwen3-4B\#best-practices} for sampling: temperature=0.7, top-p=0.8, top-K=20 for instruct models; and temperature=0.6, top-p=0.95, top-K=20 for thinking models. For open-weight baseline models, we evaluate them under the same setting and report the higher results between our replication and their reported ones.

\begin{table*}[t]
    \centering
    % \resizebox{0.95\textwidth}{!}{
    \begin{tabular}{l|c c c c c |c}
        \toprule
        \textbf{Model Name}  & \textbf{GPQA-Diamond} & \textbf{SuperGPQA} &\textbf{GPQA-General}& \textbf{HLE} & \textbf{MMLU-Pro} & \textbf{Avg}\\
        \midrule
        Qwen3-4B-Base & 36.7 & 28.5 & 5.62 & 0.92 & 50.6 & 24.5\\
        \midrule
        \multicolumn{7}{c}{\textit{\small Thinking Models}} \\
        \midrule
        o1-mini & 60.0 & 45.2 & 25.8 & 5.68 & \textbf{80.3} & 43.4\\
        QwQ-32B & 55.3 & 43.6 & 16.9 & 4.84 & 66.2 & 37.4\\ 
        Qwen3-4B thinking & 55.9 & 42.7 & 20.9& 4.52 & 70.4 & 38.9\\
        % Qwen3-4B-Thinking-2507 & 65.8 & 47.8 & 32.1& 6.84 & 74.0 & 45.3 \\
        R1-Distill-Qwen-32B & 62.1 & 39.3 & 30.9 & 5.36 & 67.5 & 41.0\\
        R1-0528-Qwen3-8B & 61.1 & 42.1 & 28.7 & 5.56 & 71.4 & 41.8\\
        \midrule
        % Dr.SCI-4B-think SFT  & 59.2 & 42.3 & 26.3& 5.40 & 67.8 & 40.5\\
        Dr.SCI-4B-think & \textbf{63.2} & \textbf{45.7} & \textbf{32.4} & \textbf{6.12} & 75.6 & \textbf{44.6}\\
        \midrule
        \multicolumn{7}{c}{\textit{\small Instruct Models}} \\
        \midrule
        GPT-4o & 50.0 & \textbf{44.4} & 22.4 & 3.48 & \textbf{74.6} & 39.0\\
        Qwen3-4B non-thinking & 41.7 & 32.0  & 9.74 & 4.44 & 58.0 & 29.2\\
        % Qwen3-4B-Instruct-2507 & 62.0 & 42.8 & 24.8 & 4.68 & 69.6 & 40.8\\
        General-Reasoner-4B  & 42.9 & 32.5 & 15.3 & 4.32  & 62.8 & 31.6\\
        General-Reasoner-Qw3-14B  & 56.1 & 39.9& 23.2  & 4.68  & 70.3 & 38.9\\
        Qwen3-4B-MegaScience &  34.9 & 33.1 & 13.2& 4.12 & 61.2 & 29.3\\
        Qwen3-8B-MegaScience &  46.5 & 38.8 & 20.4& 4.72 & 67.3 & 35.5\\
        Qwen3-14B-MegaScience &  50.5 & 44.4 & 23.6 & 4.64 & 71.9 & 39.0\\
        Qwen3-4B-VeriFree  & 42.4 & 35.1 & 16.7& 4.04 & 63.5 & 32.3\\
        Qwen3-8B-VeriFree  & 44.4 & 38.0 & 19.3 & 4.36& 67.2 & 34.6\\
        
        \midrule
        % Dr.SCI-4B-instruct SFT  & 50.6 & 39.0 & 17.8& 4.52  & 59.2 & 34.2\\
        Dr.SCI-4B-instruct & \textbf{56.6} & 43.6 & \textbf{24.3}& \textbf{5.36} & 71.0 & \textbf{40.2}\\
        
        \bottomrule

    \end{tabular}
    % }
    \caption{Full experiment results of models across scientific reasoning benchmarks. We highlight best performance for thinking and instruct models using \textbf{bold} text. Dr.\ SCI surpasses baseline methods in scientific reasoning, delivering highest overal score.}
    \label{tab: experiment all result}
    % \vspace{-5mm}
\end{table*}

\subsection{Baselines}

We compare our method against a broad set of post-trained scientific reasoning models. These include:

(i) \textbf{Qwen3-4B}~\cite{yang2025qwen3}: Official Qwen3-4B model in both thinking and non-thinking mode;

(ii) R1~\cite{deepseekai2025deepseekr1incentivizingreasoningcapability} Distill Models: \textbf{R1-Distill-Qwen-32B} and \textbf{R1-0528-Qwen3-8B};

(iii) \textbf{QwQ-32B}~\cite{qwq32b}: reasoning post-trained model of the Qwen2.5~\cite{qwen2.5} series;

(iv) Proprietary models: OpenAI's \textbf{GPT-4o}~\cite{hurst2024gpt} and \textbf{o1-mini}~\cite{o1-mini};

(v) General Reasoner~\cite{ma2025general}: \textbf{General-Reasoner-4B} and \textbf{General-Reasoner-Qw3-14B} are Qwen3= models post-trained on WebInstruct-verified~\cite{ma2025general};

(vi) MegaScience~\cite{fan2025megascience}: \textbf{Qwen3-4B-MegaScience}, \textbf{Qwen3-8B-MegaScience} and \textbf{Qwen3-14B-MegaScience} are post-trained versions of Qwen3 models on MegaScience dataset;

(vii) VeriFree~\cite{zhou2025reinforcing}: \textbf{Qwen3-4B-VeriFree} and \textbf{Qwen3-8B-VeriFree} are reasoning models post-trained on WebInstruct-verified~\cite{ma2025general} using probability-based rewards instead of verification-based rewards.

% \begin{itemize}
%     \item \textbf{Qwen3-4B}~\cite{yang2025qwen3}: The official Qwen3-4B release post-trained by Qwen team. We compare our thinking and instruct model with its "thinking" and "non-thinking" mode respectively.
%     \item \textbf{R1-Distill}: R1-0528-Qwen3-8B is distilled version of Qwen3-8B-Base~\cite{yang2025qwen3} using DeepSeek-R1-0528~\cite{deepseekai2025deepseekr1incentivizingreasoningcapability} as teacher. R1-Distill-Qwen-32B is distilled version of Qwen2.5-32B~\cite{qwen2.5} using DeepSeek-R1 as teacher.
%     \item \textbf{QwQ-32B}: QwQ-32B~\cite{qwq32b} is a reasoning model of the Qwen2.5~\cite{qwen2.5} series.
%     \item \textbf{GPT-4o}~\cite{hurst2024gpt}: We use OpenAI GPT-4o-2024-11-20 release.
%     \item \textbf{General-Reasoner}~\cite{ma2025general}: General-Reasoner-4B and General-Reasoner-Qw3-14B, reasoning models based on Qwen3-4B-Base and Qwen3-14B-Base. Both models are post-trained on WebInstruct-verified via RL with their fine-tuned GenRM.
%     \item \textbf{Qwen3-MegaScience}~\cite{fan2025megascience}: Qwen3-4B-MegaScience, Qwen3-8B-MegaScience and Qwen3-14B-MegaScience are post-trained versions of Qwen3-4B, Qwen3-8B, Qwen3-14B models by MegaScience.
%     \item \textbf{VeriFree}~\cite{zhou2025reinforcing}: Qwen3-4B-VeriFree and Qwen3-8B-VeriFree are reasoning models post-trained on WebInstruct-verified using probability-based rewards instead of verification-based rewards from Qwen3-4B-Base and Qwen3-8B-Base respectively.
% \end{itemize}

\subsection{Experiment Results}
Table~\ref{tab: experiment all result} summarizes the main results on scientific reasoning benchmarks. Overall, Dr.\ SCI yields substantial improvements over the base model in both thinking and instruct modes, demonstrating that our post-training pipeline markedly strengthens scientific reasoning capabilities. The gains are particularly pronounced on open-ended evaluation. On GPQA-General, Dr.\ SCI-4B-think achieves a score of 32.4 and Dr.\ SCI-4B-instruct achieves 24.3, compared to 5.62 for the base model. These results represent large absolute improvements and rank the best among thinking and instruct models at comparable scale.

Across all benchmarks, our 4B models consistently outperform strong post-trained baselines with the same backbone, and in many cases, surpass larger models up to 32B parameters. Notably, Dr.\ SCI-4B-think exceeds the proprietary o1-mini on GPQA-Diamond, SuperGPQA, and HLE, while Dr.\ SCI-4B-instruct outperforms GPT-4o on GPQA-Diamond, GPQA-General, and HLE. These results indicate that the improvements achieved by Dr.\ SCI go beyond what can be attributed to model scale alone.

Taken together, the results show that Dr.\ SCI substantially enhances scientific reasoning capabilities, with especially strong gains in open-ended settings where answers must be evaluated beyond rule-based verification. This highlights the effectiveness of our data processing pipeline, curriculum design, and rubric-guided reinforcement learning in addressing the core challenges of open-ended scientific reasoning.
\subsection{Analysis}

We conduct ablation studies to isolate and quantify the contribution of each component in Dr.\ SCI, including \textbf{Exploration-Expanding SFT}, the \textbf{Dynamic Difficulty Curriculum}, and \textbf{SciRubric-Guided RL}. All experiments follow the setup in Section~\ref{sec: experiments implementation} and are evaluated on GPQA-Diamond~\cite{rein2024gpqa} and GPQA-General to balance representativeness and computational efficiency.

\subsubsection{Exploration-Expanding SFT}

\label{sec: ablation data selection}
We ablate the effectiveness of Exploration-Expanding SFT (EESFT) against two baselines: \textbf{ZeroRL}, which applies RL directly to  base model, and \textbf{SFT+RL}, which uses random SFT data from $\mathcal{D}_{\text{think}}$ and $\mathcal{D}_{\text{inst}}$ (Section~\ref{sec: method 4gram}). All methods share an identical RL stage using same 100K verifiable questions from Dr.\ SCI. Results are summarized in Table~\ref{tab: analysis coverage}.

% We conduct a comprehensive ablation study to validate the effectiveness of our Exploration-Expanding SFT (EESFT). We compared against two baselines: \textbf{ZeroRL}, where we initialize RL directly from the base model without any SFT; and \textbf{SFT+RL}, where we use uniformly random-sampled SFT data from our data pools $\mathcal{D}_{think}$ and $\mathcal{D}_{inst}$ (introduced in Section~\ref{sec: method 4gram}). For all SFT methods, we apply an identical RL stage using the same 100K randomly selected verifiable questions from \textsc{SciReason}. Table~\ref{tab: analysis coverage} summarizes the results across both thinking and instruction modes, we report both performance of SFT checkpoint (no background) and performance gain from RL (\gainexp{blue} background for ours, \gainexpg{grey} for baseline).

\begin{table}[t]
    \centering
    \resizebox{\columnwidth}{!}{
    \begin{tabular}{l|c|c| c c}
        \toprule
        Method & \# SFT Data & \# 4-grams & GPQA-D & GPQA-G \\
        \midrule
        ZeroRL & 0 & 0 & 36.9\gaing{4.4} & 6.7\gaing{1.8} \\
        \midrule
        \multicolumn{5}{c}{\textit{Thinking Mode}}\\
        \midrule
        SFT+RL & \multirow{2}{*}{10K}& 15.66M & 42.3\gaing{1.8} &16.3\gaing{4.7} \\
        EESFT+RL & & 37.29M & 44.2\gainb{3.0} & 22.1\gainb{6.1}\\
        \midrule
        SFT+RL & \multirow{2}{*}{50K}& 78.23M & 43.8\gaing{5.2} &24.8\gaing{2.1}\\
        EESFT+RL & & 139.6M & 47.5\gainb{4.9} & 24.6\gainb{5.8} \\
        \midrule
        EESFT+RL & 1M & 1.564B & 59.2\gainb{3.3} & 26.3\gainb{6.5}\\
        \midrule
        \multicolumn{5}{c}{\textit{Instruct Mode}}\\
        \midrule
        SFT+RL & \multirow{2}{*}{50K}& 24.65M & 43.6\gaing{2.5} &11.3\gaing{3.4} \\
        EESFT+RL & & 38.81M & 45.2\gainb{3.2} & 12.6\gainb{3.8}\\
        \midrule
        SFT+RL & \multirow{2}{*}{250K}& 122.0M& 44.3\gaing{2.6} & 12.9\gaing{3.1}\\
        EESFT+RL & & 142.3M & 46.2\gainb{4.1} & 14.5\gainb{4.2}\\
        \midrule
        EESFT+RL & 1M & 488.7M & 50.6\gainb{3.9} &17.8\gainb{4.9} \\
        \bottomrule
    \end{tabular}
    }
    \caption{Ablation Study of Exploration Expanding SFT. EESFT improves both SFT performance (white) and RL growth (\gainexp{highlight}) due to enhanced exploration compared to standard baselines.}
    \label{tab: analysis coverage}
    % \vspace{-8mm}
\end{table}
Across all dataset sizes and both thinking and instruct modes, EESFT consistently yields stronger SFT checkpoints and, more importantly, substantially larger performance gains after RL. 
% For example, with 50K thinking-mode SFT data, EESFT achieves 47.5\% on GPQA-Diamond after SFT and improves by +4.9\% during RL, compared to 44.7\% and +2.6\% for standard SFT. Similar trends hold on GPQA-General and in instruct mode, indicating that EESFT reliably amplifies downstream RL effectiveness.
We attribute these gains to increased reasoning-pattern coverage during SFT. As shown in Table~\ref{tab: analysis coverage}, EESFT selects data with significantly more unique 4-grams than random sampling (e.g., 139.6M vs.\ 78.23M at 50K in thinking mode). This expanded coverage provides a broader exploration space prior to RL, leading to stronger initial policies and larger performance improvements during optimization.

Scaling EESFT further strengthens this effect. As SFT size increases to 1M, it continues to accumulate more unique 4-grams (1.564B in thinking mode and 488.7M in instruct mode), accompanied by the highest final performance. These results support the role of our EESFT in raising the exploration ceiling and unlocking larger RL gains.
% At all dataset sizes, our Exploration-Expanding SFT consistently achieves higher SFT performance and larger RL performance gains. For example, with 50K training samples in thinking mode, our method attains 47.5\% on GPQA-diamond after SFT and gains +4.9\% during RL, substantially exceeding standard SFT's 44.7\% and +2.6\% improvement. Similar gaps hold for GPQA-general and across instruct mode, demonstrating the robustness of our method.

% A key advantage behind this success is improved reasoning pattern diversity during SFT, i.e. the number of unique 4-grams in training data as shown in Table~\ref{tab: analysis coverage}. For instance, our 50K selection in thinking mode yields 139.6M novel 4-grams versus 78.23M for standard SFT. This increased coverage of science reasoning patterns directly correlates with superior performance: higher coverage during SFT not only start RL from a stronger policy but also exhibit amplified performance growth during RL due to better exploration potentials.

% Scaling Exploration-Expanding SFT further accentuates these benefits. As we increase to 1M samples, our method steadily accumulates more novel 4-grams (1.564B in thinking mode, 488.7M for instruct mode), delivering even higher performances. This scaling effect confirms that maximizing coverage of reasoning patterns during SFT is crucial for unlocking larger overall gains throughout post training, with our 1M-example models achieving highest performances.

\subsubsection{Dynamic Difficulty Curriculum}
We evaluate the effectiveness and efficiency of our dynamic difficulty curriculum for RL. All runs initialize from the same 250K instruct-mode EESFT checkpoint. We compare against three baselines trained with 100K verifiable questions per epoch: \textbf{Random} (uniform sampling), \textbf{No Easy} (difficulty $0/8$--$6/8$), and \textbf{Hard Only} (difficulty $0/8$--$4/8$).

Our curriculum maintains a small, adaptive training set by replacing mastered questions with harder ones over time. We evaluate two variants: (i) a \textbf{100K pool} variant that trains on only 13.1K data per epoch (86.9\% less per-epoch compute), and (ii) a \textbf{461K pool} variant that uses 82.4K examples per epoch, matching the compute budget of the baselines.

% We demonstrate the effectiveness and efficiency of our dynamic difficulty curriculum for RL training. All runs initialize from the 250K instruct mode Exploration-Expanding SFT checkpoint. We compare against three baselines trained on 100K verifiable questions per epoch: \textbf{Random} (uniform sampling), \textbf{No Easy} (difficulty $0/8$--$6/8$), and \textbf{Hard Only} (difficulty $0/8$--$4/8$).

% Our curriculum trains on only a per-epoch subset that is continuously re-sampled: mastered questions are replaced with fresh, harder ones are added between epochs.
% Thus we include two variants: (i) \textbf{100K pool} with only 13.1K examples per epoch (86.9\% less per-epoch compute), and (ii) \textbf{461K pool} with 82.4K per epoch (comparable compute to baselines).

\begin{table}[t]
    \centering
    \resizebox{\columnwidth}{!}{
    \begin{tabular}{l|c c| c c}
        \toprule
        RL Data & \# Data & \# Data / Ep. & GPQA-D & GPQA-G \\
        \midrule
        Random  & 100K & 100K & 50.3 & 18.7 \\
        No Easy  & 100K & 100K & 52.4 &  19.2\\
        Hard Only & 100K & 100K & 51.7 &  17.4\\
        \midrule
        Our Curr.& 100K & 13.1K & 51.9 &  18.3\\
        Our Curr. Long& 461K & 82.4K & \textbf{54.2} & \textbf{20.6} \\
        \bottomrule
    \end{tabular}
    }
    \caption{Ablation of our dynamic difficulty curriculum. Our curriculum improves RL efficiency and effectiveness at the same time compared to static filtering methods.}
    \label{tab: analysis difficulty}
    % \vspace{-7mm}
\end{table}

\begin{figure}[h]
    \centering

    \begin{subfigure}[t]{0.95\columnwidth}
    \centering 
    \includegraphics[width=0.95\columnwidth]{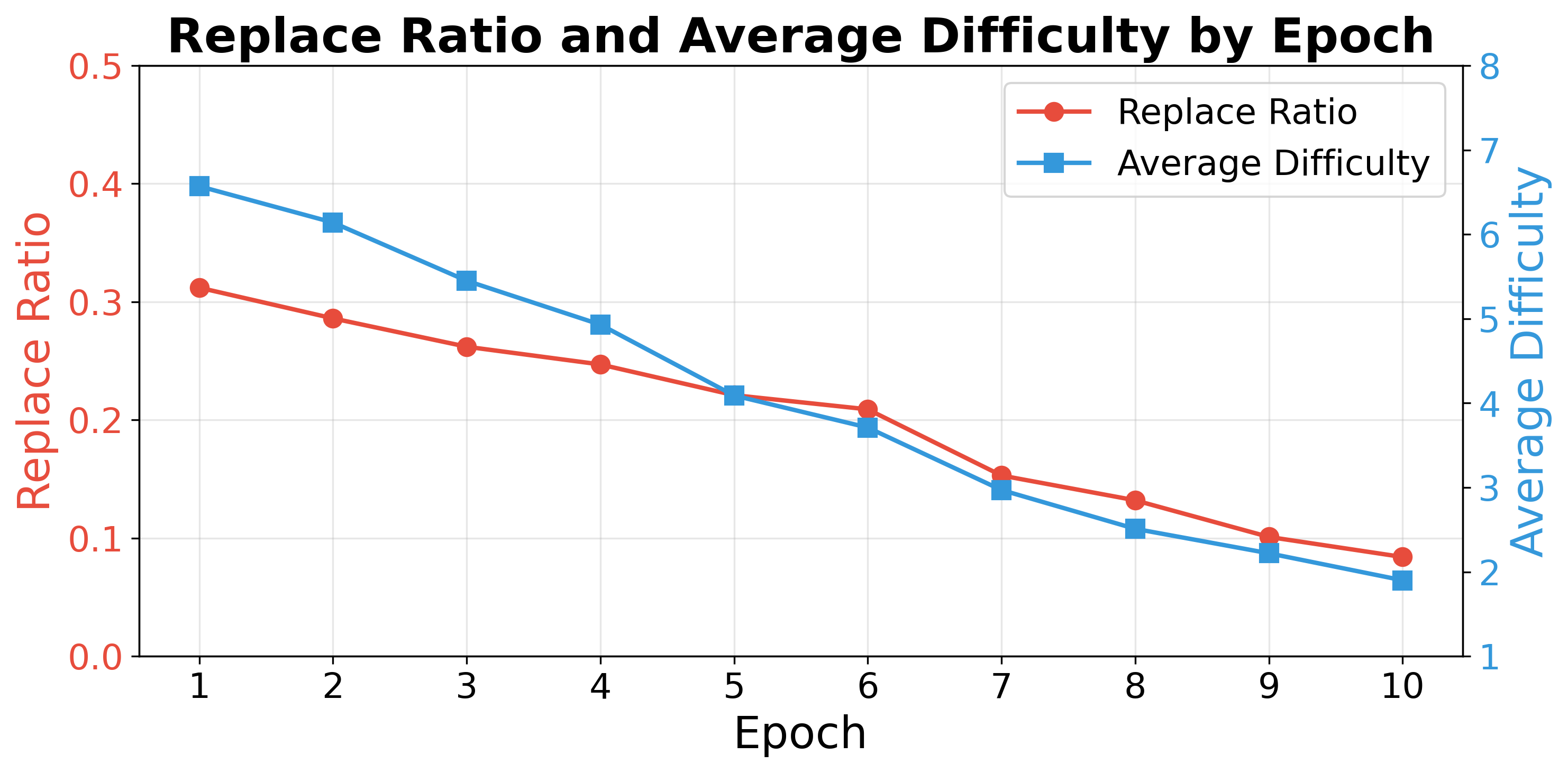}
    \caption{Replace Ratio and Average Difficulty for RL data.} 
    \label{fig: analysis difficulty replace ratio}
    \end{subfigure} 

    \begin{subfigure}[t]{0.95\columnwidth}
    \centering 
    \includegraphics[width=0.95\columnwidth]{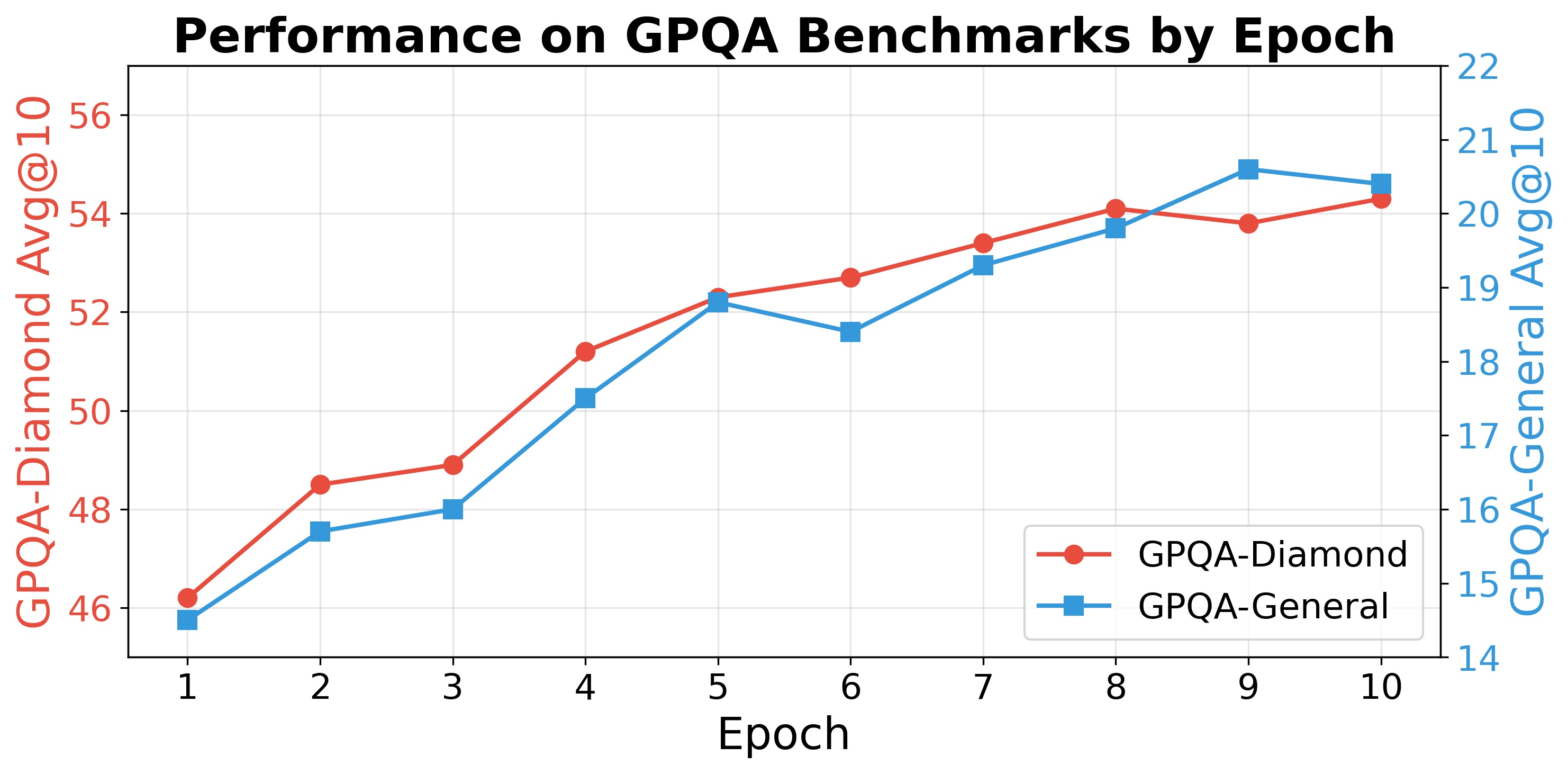}
    \caption{Performance Growth throughout RL training.
    } 
    \label{fig: analysis difficulty performance}
    \end{subfigure} 
    
    \caption{Dynamics and performance of the dynamic difficulty curriculum. \textbf{(a)} Our curriculum dynamically adjusts the average difficulty of training data accoring to current model capabilities. \textbf{(b)} This yields steady performance growth in scientific reasoning.}
    \label{fig: analysis difficulty}
    % \vspace{-6mm}
\end{figure}

As shown in Table~\ref{tab: analysis difficulty}, our curriculum achieves a favorable balance between performance and efficiency. Despite using only 13.1K examples per epoch, the compute-efficient variant outperforms \textbf{Random} sampling and matches the performance of \textbf{No Easy} and \textbf{Hard Only}, which require the full 100K examples per epoch. Moreover, when scaling the pool size while keeping per-epoch compute comparable, the curriculum yields clear gains over all baselines. Figure~\ref{fig: analysis difficulty} further illustrates how the curriculum automatically shifts training toward harder data over epochs, alongside steady performance improvements.

These results indicate that dynamically targeting samples with appropriate difficulty preserves informative reward signals and leads to both higher accuracy and improved compute efficiency. By continuously matching training difficulty to the model’s capability, the curriculum enables more effective and scalable RL training.

% As shown in Table~\ref{tab: analysis difficulty}, the curriculum attains better or comparable performance while using far fewer samples per epoch in the compute-efficient setting, and yields the best final scores when scaling the total pool while keeping per-epoch compute similar. Figure~\ref{fig: analysis difficulty} further illustrates how the curriculum automatically shifts training toward harder data over epochs, alongside steady improvements on GPQA-diamond and GPQA-general benchmarks.

% As shown in Table~\ref{tab: analysis difficulty}, our dynamic difficulty curriculum emphasizes both performance and data efficiency. Training on merely 13.1K dynamically selected data per epoch, it delivers better performance than \textbf{Random}, and comparable RL performance than pre-RL filtering strategies \textbf{No Easy}, \textbf{Hard Only} that utilizes the full 100K each epoch.  

% Moreover, When we scale the total data but keep per-epoch compute comparable,  the curriculum raises final RL scores by a clear margin. We plot statistics of our curriculum as well as the performance growth in Figure~\ref{fig: analysis difficulty}. These results confirm that dynamically targeting the most instructive questions for the current policy simultaneously improves accuracy and compute efficiency, enabling large-scale RL without extra training cost.

\subsubsection{SciRubric-Guided RL}
\label{sec: ablation Open-Ended}
We ablate reward designs for RL on 100K \textit{open-ended} questions from Dr.\ SCI. We compare against two baselines: \textbf{GenRM}, which assigns binary rewards using a generative reward model, and \textbf{RaR}~\citep{gunjal2025rubrics}, which aggregates rubric satisfaction via weighted averaging. All runs initialize from the same 50K thinking-mode EESFT checkpoint. We further evaluate a \emph{unified} RL setting that combines rule-based rewards on verifiable questions with our rubric-guided rewards on open-ended questions, mirroring the full Dr.\ SCI pipeline.
% We ablate reward designs for RL on 100K \textit{open-ended} questions from \textbf{Dr.\ SCI}. We compare against two baselines: \textbf{GenRM}, which assigns a binary reward using a generative reward model with the prompt in the Appendix; and \textbf{RaR}~\citep{gunjal2025rubrics}, which aggregates rubric satisfaction scores via a weighted average. All runs initialize from the 50K thinking mode Exploration-Expanding SFT checkpoint. We further evaluate a \emph{unified} RL setting that combines rule-based rewards on verifiable questions (RLVR) with our rubric-based rewards on open-ended questions, mirroring the final Dr.\ SCI pipeline. We report GPQA-Diamond and GPQA-General (GPQA-G; see Section~\ref{sec: experiment evaluation}).

As shown in Table~\ref{tab: analysis rubric}, GenRM leads to training collapse and underperforms the initial policy on both benchmarks, consistent with reward hacking and spurious positive feedback. RaR yields modest improvements but is limited by rigid score aggregation, which empirically encourages partial-credit accumulation and overly long responses rather than correct problem solving (see Appendix~\ref{appn: example rewards} for qualitative examples and anlaysis).

% As shown in Table~\ref{tab: analysis rubric}, \textbf{GenRM} leads to training collapse, underperforming the initial policy on both benchmarks, consistent with reward hacking and spurious positive feedback that degrades reasoning quality. \textbf{RaR} yields moderate gains but is limited by rigid score aggregation; empirically, it often incentivizes overly long reasoning and final answers that attempt to accumulate partial credit rather than solving the problem \footnote{Qualitative examples are provided in the Appendix.}. 

In contrast, \textbf{SciRubric-Guided RL} delivers consistent gains, with particularly strong improvements on the open-ended benchmark GPQA-General. Moreover, the \emph{unified} training regime that jointly optimizes verifiable questions and open-ended questions achieves the best overall performance, outperforming both RLVR-only and open-ended-only training. These results demonstrate the necessity of structured, correctness-aware rewards for open-ended scientific reasoning and highlight the benefit of unifying verifiable and open-ended supervision within a single RL framework.

\begin{table}[t]
    \centering
    \resizebox{\columnwidth}{!}{
    \begin{tabular}{l|c| c c}
        \toprule
        Reward Type & \# Data & GPQA-D & GPQA-G \\
        \midrule
        Init Policy & - & 47.5 & 16.3 \\
        RLVR & 100K RLVR & 52.4 & 20.8 \\
        \midrule
        GenRM & 100K Open & 42.7 & 9.8  \\
        RaR & 100K Open & 48.7 & 18.7  \\
        \midrule
        Our SciRubric & 100K Open & 50.8 & 23.6 \\
        Our Unified & 100K RLVR + 53K Open & 54.2 & 24.9 \\
        \bottomrule
    \end{tabular}
    }
    \caption{Ablation of SciRubric-Guided RL. Our SciRubric reward enables stable performance growth on open-ended RL.}
    \label{tab: analysis rubric}
    % \vspace{-8mm}
\end{table}

%%%%%%%%%%%%%%%%%%%%%%%%%%%%%%%%%%%%%%%%%%%%%%%%%%%%%%%%%%%%%%%%%%%%%%%%%%%%%%%
%%%%%%%%%%%%%%%%%%%%%%%%%%%%%%%%%%%%%%%%%%%%%%%%%%%%%%%%%%%%%%%%%%%%%%%%%%%%%%%
% \vspace{-3mm}
\section{Conclusion}
% \vspace{-2mm}
We introduce \textbf{Dr.\ SCI} as a principled foundation for scientific reasoning post-training, addressing key practical bottlenecks that have hindered progress in this domain: fragile data curation, poorly calibrated difficulty, and unreliable evaluation for open-ended scientific questions. Dr.\ SCI combines a scalable data processing pipeline with fine-grained supervision and quality control; with a unified post-training framework integrating Exploration-Expanding SFT, a Dynamic Difficulty Curriculum, and SciRubric-Guided RL, enabling stable and effective reinforcement learning across both verifiable and open-ended settings.

Empirically, Dr.\ SCI delivers substantial gains from a compact 4B backbone model and consistently surpasses strong post-trained baselines, including much larger and proprietary models. These results demonstrate that principled data processing and correctness-aware rewards are critical for advancing open-ended scientific reasoning, and provide a practical recipe for future research in this underexplored yet increasingly important area.
% In this paper, we introduce Dr.\ SCI as a foundation for scientific reasoning post-training, addressing the practical bottlenecks that have limited progress: imperfect data curation, unchecked difficulty, and unreliable evaluation for open-ended science questions.
% Dr.\ SCI delivers (i) a 1M scientific reasoning dataset across eight STEM subjects with explicit verifiable/open-ended splits, scalable difficulty annotations, and fine-grained rubrics for open-ended evaluation, and (ii) a unified post-training pipeline including: \textbf{Exploration-Expanding SFT}, \textbf{Dynamic Difficulty Curriculum}, and \textbf{SciRubric-Guided RL}, that enables stable, scalable RL over both rule-checkable and open-ended supervision.
% Across diverse benchmarks, Dr.\ SCI yields substantial gains from a compact 4B backbone and surpasses strong post-trained baselines, demonstrating that principled data processing and rubric-guided rewards can unlock scale-defying improvements in scientific reasoning, and providing a practical recipe to accelerate future work in this underexplored area. 
%%%%%%%%%%%%%%%%%%%%%%%%%%%%%%%%%%%%%%%%%%%%%%%%%%%%%%%%%%%%%%%%%%%%%%%%%%%%%%%
%%%%%%%%%%%%%%%%%%%%%%%%%%%%%%%%%%%%%%%%%%%%%%%%%%%%%%%%%%%%%%%%%%%%%%%%%%%%%%%
\section*{Impact Statement}
This paper presents work whose goal is to advance the field of machine learning, with a particular focus on improving post-training methodologies for scientific reasoning in large language models.

\textbf{Potential Benefits}
By introducing a principled data processing pipeline, stable curriculum design, and correctness-aware reward mechanisms for open-ended scientific questions, this work may contribute to more reliable and interpretable scientific reasoning capabilities in future language models. Such improvements could support downstream applications in education, scientific research assistance, and knowledge-intensive domains, where structured reasoning and faithful explanations are critical.

\textbf{Limitations and Risks.}
At the same time, models equipped with stronger scientific reasoning abilities may be misused to generate plausible-sounding but incorrect scientific explanations if deployed without appropriate safeguards or human oversight. While our work focuses on improving training stability and evaluation reliability, it does not solve broader challenges related to factuality, misuse, or overreliance on automated scientific advice.

\textbf{Ethical Considerations.}
Our dataset is constructed exclusively from publicly available sources, and the proposed methods operate at the level of model training rather than direct deployment. We do not foresee novel ethical risks beyond those commonly associated with large language models trained for reasoning tasks. Nevertheless, we emphasize that responsible use of such models requires careful deployment practices, transparency about model limitations, and continued human involvement in high-stakes scientific decision-making.

Overall, we believe this work represents a methodological advance in scientific reasoning post-training, with ethical implications that are largely aligned with existing discussions in the machine learning community, and no immediate societal risks that warrant special concern beyond established best practices.

\bibliography{example_paper}
\bibliographystyle{icml2026}

%%%%%%%%%%%%%%%%%%%%%%%%%%%%%%%%%%%%%%%%%%%%%%%%%%%%%%%%%%%%%%%%%%%%%%%%%%%%%%%
%%%%%%%%%%%%%%%%%%%%%%%%%%%%%%%%%%%%%%%%%%%%%%%%%%%%%%%%%%%%%%%%%%%%%%%%%%%%%%%
% APPENDIX
%%%%%%%%%%%%%%%%%%%%%%%%%%%%%%%%%%%%%%%%%%%%%%%%%%%%%%%%%%%%%%%%%%%%%%%%%%%%%%%
%%%%%%%%%%%%%%%%%%%%%%%%%%%%%%%%%%%%%%%%%%%%%%%%%%%%%%%%%%%%%%%%%%%%%%%%%%%%%%%
\newpage
\appendix
\onecolumn
%%%%%%%%%%%%%%%%%%%%%%%%%%%%%%%%%%%%%%%%%%%%%%%%%%%%%%%%%%%%%%%%%%%%%%%%%%%%%%%
%%%%%%%%%%%%%%%%%%%%%%%%%%%%%%%%%%%%%%%%%%%%%%%%%%%%%%%%%%%%%%%%%%%%%%%%%%%%%%%
\section{Related Works}
\subsection{Scientific Reasoning Dataset}
Recent efforts have begun to scale up open resources for training scientific (and broader) reasoning models, but they differ substantially in supervision structure and their suitability for RL-centric post-training. 

Distilled SFT corpora such as OpenThoughts~\cite{guha2025openthoughts} and NVIDIA Nemotron releases~\citep{NemotronPostTrainingDatasetV1,bercovich2025llamanemotronefficientreasoningmodels,Nemotron_Cascade_Scaling_Cascaded_Reinforcement_Learning} primarily provide teacher-generated responses (often with long reasoning traces) in an instruction-format, and are explicitly used to train strong reasoners via supervised fine-tuning. For example, OpenThoughts~\cite{guha2025openthoughts} reports training its strongest open-data models using only SFT, and its pipeline expands data largely by sampling multiple teacher responses per prompt, i.e. 6k science questions with 16 responses each. Similarly, Nemotron-Science~\cite{NemotronPostTrainingDatasetV1} is released as chat-style message pairs containing an assistant solution (and optionally a reasoning field), but it does not provide a separate, canonical reference-answer/verification interface intended for automated reward computation. As a result, these datasets are well-suited for SFT bootstrapping, yet are less ideal when RL is the core optimization stage: they typically lack instance-level verification signals and do not standardize evaluation targets in a way that supports stable reward design across heterogeneous scientific questions.

A complementary line of work constructs large-scale science mixtures with reference answers to support RL training, including MegaScience~\cite{fan2025megascience} and its textbook-derived component TextbookReasoning, NaturalReasoning~\cite{yuan2025naturalreasoning}, WebInstruct-verified~\cite{ma2025general} etc. MegaScience\cite{fan2025megascience} curates a science reasoning dataset TextbookReasoning based on college level textbooks, and aggregates multiple public science sources to form a dataset of 1.25M samples with reference answers. However, their dataset is not suitable for direct post-training due to lack of quality control: MegaScience contains over easy questions like ``Change 1,929 meters to kilometers." and malformed reference answers like ``\$\textbackslash boxed\{\textbackslash begin\{aligned\}\textbackslash text\{Metric variation:\} ...\textbackslash text\{ is the Christoffel connection.\} \textbackslash end\{aligned\}\}\$", to name a few. NaturalReasoning~\cite{yuan2025naturalreasoning} scales to 2.8M questions spanning many domains including STEM by generating questions from pretraining corpora and extracting reference answers when possible. Yet, in practice these resources still leave key post-training metadata under-specified for RL: a non-trivial fraction of instances may lack reference answers (210k among 1.15 samples open sourced by natural reasoning), and many reference answers are long natural-language texts and unprocessed, making correctness difficult to validate via simple rules or naive matching. 

To move forward, WebInstruct-verified~\cite{ma2025general} provides a curated set of short reference answer questions across disciplines, it focuses on rule-checkable supervision and offers a finetuned reward model capable of assessing responses during RL for open-ended questions. But model-based rewards are natively vulnerable to reward hacking~\citep{gao2023scaling,zhou2025reinforcing,gunjal2025rubrics}, thereby leading to suboptimal post-training results. Recently, RaR-Science-20k ~\cite{gunjal2025rubrics} explore rubric-based methods for scientific post-training, they pair 20k curated science problems with generated rubrics used for evaluation during train time. But the scale of dataset remains relatively small and in practice they rubrics often fail to verify the accuracy of final answer provided by model.

These limitations motivate us to construct a large scale, high quality, RL-ready scientific reasoning resource. Our Dr.\ SCI dataset includes 1M challenging scientific reasoning questions through systematically curation of open-source science data with explicit verifiable/open-ended splits, scalable difficulty annotations, and fine-grained rubrics that operationalize evaluation for open-ended questions, enabling stable RL beyond strictly rule-verifiable settings.

\subsection{Scientific Reasoning Post-training}
A growing line of work applies RL to improve LLM's capability of scientific reasoning beyond SFT. Early successes such as R1~\cite{deepseekai2025deepseekr1incentivizingreasoningcapability} and GLM-4.5~\cite{5team2025glm45agenticreasoningcoding} leverages rule-based rewards for science RL, and utilizes multiple-choice questions for science domain where correctness can be reduced to selecting the right option.  While effective when verification is straightforward, this paradigm struggles to scale to the broader landscape of scientific reasoning, where answers are often free-form explanations and verification is inherently under-specified.

To extend RL with verifiable rewards (RLVR) beyond strictly structured tasks, later work attempts to model-based verifiers or reward models for science domains. \citet{su2025crossing} study RLVR across diverse domains (including science-related areas) and propose using model-based scoring to handle less structured reference answers, demonstrating that RL can be driven by learned verification signals when expert-written references exist. Similarly, \citet{ma2025general} introduce General-Reasoner, which trains a generative reward model to support broader answer formats and enables RL beyond narrow rule-based checking. However, model-based rewards are vulnerable to reward hacking and spurious reward correlations, since the reward model itself becomes an optimization target and may be exploited by the policy~\citep{gao2023scaling,zhou2025reinforcing,gunjal2025rubrics}.

To reduce reliance on external verifiers, \citet{zhou2025reinforcing} propose VeriFree, which bypasses explicit verification and instead optimizes the policy to maximize the probability of the reference answer under the model. This design removes the need to maintain a separate verifier during training, but its reward signal can still be noisy and inaccurate for open-ended scientific questions, where reference answers are long natural-language explanations and correctness is not well captured by likelihood of a single reference. 

More recently, rubric-based RL aims to address open-ended evaluation by decomposing quality into structured criteria. RaR~\cite{gunjal2025rubrics} using checklist-style rubrics generated by strong LLMs to produce more interpretable reward signals and reporting gains in science and medical settings over LLM-as-judge baselines. \citet{huang2025reinforcement} further introduce Reinforcement Learning with Rubric Anchors, scaling rubric reward systems to large scale open-ended tasks and general reasoning with structured carefully curated scoring systems, but results in marginal performance on general capabilities and even degraded performance on scientific and reasoning benchmarks like GPQA-diamond. In practice, we found existing rubric-based methods suboptimal for scientific reasoning when reward aggregation does not sufficiently enforce final-answer correctness, leading to pathological behaviors such as overly long responses or ``point-chasing'' that maximizes rubric scores without solving the problem.

These gaps in current scientific reasoning motivate our approach: Dr.\ SCI jointly redesigns data, curriculum, and reward for scientific RL by providing RL-ready open-ended supervision with fine-grained rubrics and explicit answer checks, enabling stable optimization across both verifiable and open-ended scientific questions.

%%%%%%%%%%% Hyperparameters
\section{Further Implementation Details}
\label{sec: appn implementation detail}
We include further implementation details not enumerated in Section~\ref{sec: experiments implementation} here. This includes training hyperparameter for coverage inspired SFT and standard SFT baselines (Table~\ref{tab: appn sft hyperparam}), and RL hyperparmeters throughout this paper (Table~\ref{tab: appn RL hyperparam}).

\begin{table}[h]
    \centering
    \begin{tabular}{l c}

    \toprule
    Hyperparameters & Value \\
    \midrule
    Learning Rate & 5e-5 \\
    Warmup Steps & 100 \\
    LR Schedule & cosine \\
    Batch Size & 1024 \\
    Context (Instruct) & 8192  \\
    Context (Thinking) & 16384 \\
    Epochs & 4.0 \\
    \bottomrule
    \end{tabular}
    \caption{Hyperparameters for SFT}
    \label{tab: appn sft hyperparam}
\end{table}

\begin{table}[h]
    \centering
    \begin{tabular}{l c}

    \toprule
    Hyperparameters & Value \\
    \midrule
    \multicolumn{2}{c}{\textit{GRPO Algorithm}} \\
    \midrule
    Train Batch Size & 1024 \\
    Prompt Length & 2048 \\ 
    Response Length & 14336 \\
    Number of Rollout & 8 \\
    PPO Batch Size & 512 \\
    Learning Rate & 1e-6 \\
    KL Loss Coefficient & 1e-3 \\
    Clip Ratio Low & 0.2 \\
    Clip Ratio High & 0.28 \\
    Epochs & 10 \\
    \midrule
    \multicolumn{2}{c}{\textit{Dynamic Difficulty Curriculum}} \\
    \midrule
    $\tau_{\text{discard}}$ & 1.0 \\ 
    $\tau_{\text{pending}}$ & 0.625\\
    $\tau_{\text{train}}$ & 0.9\\
    \midrule
    \multicolumn{2}{c}{\textit{Weights for Rubrics}} \\
    Essential & 1.0 \\
    Important & 0.7 \\
    Optional & 0.3 \\
    Pitfall & 0.9 \\
    Final Answer & 4.0 \\    
    \bottomrule
    \end{tabular}
    \caption{Hyperparameters for RL algorithm}
    \label{tab: appn RL hyperparam}
\end{table}
We use a simple instruction for all training data and evaluation questions as shown below in Listing~\ref{lst:appn_prompt_template}. The ``<SUBJECT>" is the subject of train and test questions, with default value as ``science"; and ``<QUESTION>" is the corresponding input question.

\begin{lstlisting}[style=prompt,caption={Instruction Template},label={lst:appn_prompt_template}]
Solve the following <SUBJECT> problem step by step. The last line of your response should be of the form: `The final answer is: \textbackslash boxed\{ANSWER\}' (without quotes) where ANSWER is your answer.\\

<QUESTION>

\end{lstlisting}

%%%%%%%%%%%%% Further Experiment Results
\section{Further Experiment Results}
\begin{table*}[h]
    \centering
    % \resizebox{0.95\textwidth}{!}{
    \begin{tabular}{l|c c c c c |c}
        \toprule
        \textbf{Model Name}  & \textbf{GPQA-Diamond} & \textbf{SuperGPQA} &\textbf{GPQA-General}& \textbf{HLE} & \textbf{MMLU-Pro} & \textbf{Avg}\\
        \midrule
        Qwen3-4B-Base & 36.7 & 28.5 & 5.62 & 0.92 & 50.6 & 24.5\\
        \midrule
        \multicolumn{7}{c}{\textit{Thinking Models}} \\
        \midrule
        o1-mini & 60.0 & 45.2 & 25.8 & 5.68 & \textbf{80.3} & 43.4\\
        \midrule
        Dr.SCI-4B-think SFT  & 59.2 & 42.3 & 26.3& 5.40 & 67.8 & 40.5\\
        Dr.SCI-4B-think & \textbf{63.2} & \textbf{45.7} & \textbf{32.4} & \textbf{6.12} & 75.6 & \textbf{44.6}\\
        \midrule
        \multicolumn{7}{c}{\textit{Instruct Models}} \\
        \midrule
        GPT-4o & 50.0 & \textbf{44.4} & 22.4 & 3.48 & \textbf{74.6} & 39.0\\
         \midrule
        Dr.SCI-4B-instruct SFT & 50.6 & 39.0 & 17.8& 4.52  & 59.2 & 34.2\\
        Dr.SCI-4B-instruct & \textbf{56.6} & 43.6 & \textbf{24.3}& \textbf{5.36} & 71.0 & \textbf{40.2}\\
        
        \bottomrule
    \end{tabular}
    % }
    \caption{Futher Experiment Results. Although the SFT checkpoints of Dr.\ SCI demonstrates strong performance, significant growth arouse during RL stage. This justifies the effectiveness of Dr.\ SCI as a whole.}
    \label{tab: appn experiment more result}
\end{table*}

The overall performance growth of Dr.\ SCI is significant, as we've shown in Table~\ref{tab: experiment all result}. Here, we provide further performance of SFT checkpoints for Dr.\ SCI-4B-think and Dr.\ SCI-4B-instruct along with representative baselines, so as to demonstrate the performance growth of Dr.\ SCI is not only due to distillation of strong teacher models like DeepSeek-R1~\cite{deepseekai2025deepseekr1incentivizingreasoningcapability} and GLM-4.6~\cite{5team2025glm45agenticreasoningcoding}. As can be seen from Table~\ref{tab: appn experiment more result}, our Exploration Expanding SFT checkpoint yields strong performance growth upon base model, with an average of 16.0 for thinking mode and 9.7 for instruct mode; the subsequent RL stage further improved the performance to exceed best performing baselines.

%%%%%%%%%%%%% GPQA-General Construction
\section{GPQA-General Construction}
\label{appn: gpqa general construct}
We construction an open-ended scientific reasoning benchmark GPQA-general from GPQA-diamond~\cite{rein2024gpqa} using GPT-4o~\cite{hurst2024gpt}. We detail the procedure below.

For each question in GPQA-diamond, we first let GPT-4o to classify if it has only one correct answer (Type 1), or answers that meets certain criteria shall all be considered correct (Type 2). We gave an example of the second type of question in GPQA-diamond in Listing~\ref{lst:data_example_gpqa_d}. As can be seen from the explaination of the example, any answer significantly greater than $10^{-7}eV$ should be considered correct. Among the provided choices, only one meets the criteria. For Type 1 questions, the reference answer is the correct choice.  For Type 2 questions, the reference answer is the criteria GPT-4o extracts from the explainations. We then prompt GPT-4o to double check if each correct choice meets the criteria it extracts before. This leads to reliable reference answers for GPQA-general benchmark. Finally we use GPT-4o to rewrite the original multiple choice question into an open-ended format. This yields our GPQA-general benchmark we use in evaluations in Section~\ref{sec: experiment evaluation}.

\begin{PromptIO}[Example Type 2 Question in GPQA-diamond: more than one answer can be considered correct.]{lst:data_example_gpqa_d}
[Original Question]
Two quantum states with energies E1 and E2 have a lifetime of 10^-9 sec and 10^-8 sec, respectively. We want to clearly distinguish these two energy levels. Which one of the following options could be their energy difference so that they be clearly resolved?

[Original Choices]
Correct: 10^-4 ev
Wrong No.1: 10^-8 ev
Wrong No.2: 10^-9 ev
Wrong No.3: 10^-11 ev

[Explaination]
According to uncertainty principle, Delta E* Delta t=hbar/2. Delta t is the lifetime and Delta E is the width of the energy level . With Delta t=10^-9 s==> Delta E1= 3.3 *10^-7 ev. And Delta t=10^-11 s gives Delta E2=3.3*10^-8 ev.
Therefore, the energy difference between the two states must be significantly greater than 10^-7 ev. So the answer is 10^-4 ev.

\end{PromptIO}

%%%%%%%%%%%%% data example
\section{Examples of Dr.\ SCI dataset}
\label{appn: example drsci}
As introduced in Section~\ref{sec: dataset}, our Dr.\ SCI dataset contains 1 million challenging scientific reasoning questions, paired with reference answer, subject category, difficulty annotation, verification split (verifiable or open-ended), and a set of fine-grained rubric to aid verification for open-ended instances. We provide 1 example of verifiable data in Listing~\ref{lst:data_example_verifiable} and 1 example of open-ended data in Listing~\ref{lst:data_example_open}.

\begin{PromptIO}[Example Verifiable Data of Dr.\ SCI]{lst:data_example_verifiable}
[Question]
A quasar is thought to be powered by the accretion of matter onto a supermassive black hole. If a object of rest mass $m$ falls onto a black hole of mass $M_{BH}$, what is the efficiency of the process in terms of the energy radiated away as a fraction of the rest mass energy of the object? Assume that the object thermalizes at the innermost stable circular orbit of the black hole and that the kinetic energy is split between emitted radiation and the kinetic energy of its orbit. Use the Newtonian approximation to estimate the emitted radiation energy and derive the efficiency $\epsilon$.

[Reference Answer]
$\epsilon \sim \frac{1}{12}$

[Subject]
physics

[Difficulty]
3/8

[Verification]
Verifiable

\end{PromptIO}

\begin{PromptIO}[Example Open-ended Data of Dr.\ SCI]{lst:data_example_open}
[Question]
What are the plesiomorphies of archosaurs, and how do they distinguish this group from other reptiles? Please provide a detailed explanation of the characteristics that are unique to archosaurs, including the presence of four-chambered hearts and pneumonic bones. Be sure to discuss the advantages and disadvantages of these characteristics and how they relate to the evolution of archosaurs.

[Reference Answer]
The plesiomorphies of archosaurs include the presence of four-chambered hearts and pneumonic bones. These characteristics distinguish archosaurs from other reptiles and provide advantages such as improved respiration and reduced body weight. The presence of air-sacs in pneumatisized bones is also a possible primitive-trait for archosaurs, which may have evolved for respiration and later became useful for other purposes such as flight and buoyancy.

[Subject]
biology

[Difficulty]
0/8

[Verification]
Open-Ended

[Rubrics]
Define Plesiomorph (Essential):
Explicitly defines a plesiomorphy as a primitive (ancestral) character state and explains why the cited archosaur traits are classified as such within the clade's evolutionary context.

Four-Chamber Heart (Essential):
States that archosaurs possess a fully divided four-chambered heart that separates oxygenated and de-oxygenated blood, contrasting it with the typical three-chambered heart of most other reptiles.

Pneumatic Bones (Essential):
Describes the presence of bone pneumatization (air-filled cavities linked to pulmonary air sacs) in archosaurs and notes that this feature is absent or rare in non-archosaur reptiles.

Comparative Distinction (Essential):
Clearly explains how the above features, plus at least one additional skeletal or soft-tissue character (e.g., antorbital fenestra, thecodont teeth, mandibular fenestra, upright gait), separate archosaurs from lepidosaurs, turtles, and other reptile groups.

Physiological Advantages (Important):
Discusses the adaptive benefits of the four-chambered heart (greater aerobic capacity, supports endothermy/diving) and pneumatic bones (weight reduction, continuous airflow) in relation to archosaur ecological success.

Potential Drawbacks (Important):
Addresses disadvantages or trade-offs, such as increased metabolic cost of a high-pressure heart or greater bone fragility/infection risk associated with pneumaticity.

Evolutionary Context (Important):
Links these traits to major evolutionary events (Triassic radiation, flight in pterosaurs and birds, crocodilian semi-aquatic lifestyle) to show how plesiomorphies facilitated later diversification.

Depth and Accuracy (Important):
Provides mechanistic or anatomical details (e.g., septum origin in heart, diverticula invading post-cranial skeleton) that demonstrate sound biological reasoning and factual correctness.

Living Examples (Optional):
Cites modern archosaurs (birds, crocodilians) to illustrate how these traits manifest today and contrasts them with representative lepidosaurs or chelonians.

Clarity and Structure (Optional):
Presents information in a well-organized, reader-friendly sequence with clear headings or logical paragraphs, enhancing readability.

Terminology Precision (Pitfall):
Mislabels derived archosaur synapomorphies as plesiomorphies for all reptiles, or uses terms like 'endothermy' and 'homeothermy' interchangeably without explanation.

Heart Misconception (Pitfall):
Incorrectly claims that crocodilians retain only a three-chambered heart or that all reptiles except birds lack complete ventricular separation.

Bone Confusion (Pitfall):
States that all reptiles possess pneumatised bones or that bone air-sacs evolved solely for flight rather than as a respiratory adaptation first.

Omission Of Comparison (Pitfall):
Fails to compare archosaurs with at least one other reptile lineage, thereby not demonstrating how the traits actually distinguish the group.
\end{PromptIO}

%%%%%%%%%%%% Qualitative Examples
\section{Qualitative Examples of Rewarding Open-ended Questions}
\label{appn: example rewards}
We give a qualitative example of different reward strategies for open-ended questions in this section. For detailed prompt we use, refer to Appendix Section~\ref{appn: prompts}

For GenRM, we first identify a key vulnerability in early experiments it often makes in our case: fail to identify and mark incorrect for meaningless, placeholder like final answer such as ``\$ANSWER" etc. This quickly drives model to output only placeholder like responses, in an early experiment with 100k data (such data may not exist in Dr.\ SCI now), the model collapsed in less than 1 epoch. We provide 2 qualitative examples of how GenRM marked these responses as correct in Listing~\ref{lst:qualitative_genrm_early_1} and Listing~\ref{lst:qualitative_genrm_early_2}. This demonstrates the weakness of GenRM, and promotes us to add ``Check if the candidate answer is complete and meaningful" in GenRM's prompt as shown in Listing~\ref{lst:appn_prompt_answer_verify}. However, even if we explicitly managed to correct these vulnerability in prompt template, GenRM still isn't capable of assigning accurate rewards. We provide 1 qualitative example where GenRM assigned a reward of 0 to a already correct enough response in Listing~\ref{lst:qualitative_genrm_FN}, where the parsed final answer just missed secondary details or features such as common positions of ``leioimyoma" but is given a reward of 0; and 1 example where it assigned 1 to a wrong response in Listing~\ref{lst:qualitative_genrm_FP},  the parsed final answer states ``iron deficiency anemia", ``vitamin B12 or folate deficiency" and ``bone marrow disorders" compared to reference answer's ``Anemia", ``Blood loss", ``Chronic diseases" and ``Hemolysis", which to the best of our knowledge provides only one correct possible conditions among reference answer's four.

For RaR~\cite{gunjal2025rubrics}, rewards are computed as a weighted average over rubric-item satisfactions. In practice, we observe a failure mode: decomposed scoring encourages \emph{partial-credit gaming}, where models produce unnecessarily long responses to accumulate points across many items instead of executing the essential steps that yield the correct solution. This brings two consequences: overlong responses, starting from the same initial policy, average response length quickly grows >8192 tokens after 100 steps of training, meanwhile RLVR responses are relatively lower (about 6300 tokens) despite even larger performance improvement; and less effective rewards and advantage signals, typically a batch of rollout would receive rewards between 0.3$\sim$0.7 and centered around 0.4$\sim$0.6, this leads to smaller reward distances and thus less representative advantage signals for GRPO algorithm. We compared the mean and std of reward of 1000 random rollouts trained using RaR, GenRM, and our SciRubric-Guided RL starting from the same initial checkpoint in Table~\ref{tab: appn reward mean std}. As can be seen from the result: RaR shows the smallest std, thereby smaller difference in reward values, which leads to less informative advantage signals and thus moderate performance growth; GenRM causes reward hacking quickly, with high mean reward but degraded performance; Our SciRubric-Guided RL gives stable and informative reward, with reasonable mean reward and much larger Std, leading to effective advantage, stable RL, and best overall performance.

\begin{table}[h]
    \centering
    \begin{tabular}{l|c c | c}
        \toprule
        Reward Type & Mean & Std. & GPQA-D\\
        \midrule 
        Base Policy & - & - & 47.5 \\
        \midrule
        GenRM & 0.5817 & 0.1238 & 42.7\\
        RaR & 0.4305 & 0.0966 & 48.7\\
        \midrule
        SciRubric-Guided RL & 0.4592 & 0.2381 & \textbf{50.8}\\
        \bottomrule
    \end{tabular}
    \caption{Mean and Std of rewards from 1000 random responses trained using different reward types. GenRM features reward hacking, characterized by high mean reward but degraded performance. RaR demonstrates moderate performance growth, primarily due to small reward divergence and less effective advantage signal, as evidenced by small std in reward. Meanwhile, our SciRubric-Guided RL leads to significant performance progress, with much larger divergence in reward signals leading to more effective advantage and RL.}
    \label{tab: appn reward mean std}
\end{table}

To address this, our \textbf{SciRubric-Guided RL} augments rubric scores with a \emph{dominant} final-answer correctness term. Concretely, we assign a substantially larger weight to the answer-check component (and optionally gate the reward so that in general $R(y)>0.5$ only when the final answer is correct), while keeping rubric items to provide fine-grained feedback on explanation quality. This ties reward to actual problem solving rather than verbosity, mitigating point-chasing behavior and improving training effectiveness and stability.

%%%%%%%%%%%%
\begin{PromptIO}[GenRM Early Failed Example 1]{lst:qualitative_genrm_early_1}
[Question]
What is the function of the nauplius larva in the life cycle of crustaceans?

[Parsed Final Answer]
\$ANSWER

[Reference Answer]
The nauplius larva's primary functions include dispersal, feeding and growth, and development, bridging the gap between embryonic development and adulthood in crustaceans.

[GenRM Verification]
### Step 1: Identify the key points in the reference answer  
- The nauplius larva's primary functions include:  
- Dispersal  
- Feeding and growth  
- Development  
- It bridges the gap between embryonic development and adulthood in crustaceans.

### Step 2: Check if the candidate answer contains all these key points  \nThe candidate answer is:  
**\"The nauplius larva is a stage in the life cycle of crustaceans that allows for dispersal, feeding, and growth, and it is a transitional stage between the embryonic stage and adulthood.\"**

This answer includes:  
- Dispersal  
- Feeding and growth  
- Transitional stage between embryonic and adult stages  
All key points from the reference answer are present.
### Step 3: Verify the factual accuracy of the candidate answer  
The candidate answer accurately describes the nauplius larva as a stage in the life cycle of crustaceans that is involved in dispersal, feeding, growth, and serves as a transitional stage between the embryonic and adult stages. This is factually correct.
### Step 4: Assess whether the candidate answer adequately addresses the question  
The question asks:  
**\"What is the function of the nauplius larva in the life cycle of crustaceans?\"**

The candidate answer clearly addresses this by explaining the functions of the nauplius larva (dispersal, feeding and growth, and development as a transitional stage), and it is complete and accurate.

### Final Decision:  
Final Decision: \textbackslash\textbackslash boxed\{Yes\}
\end{PromptIO}

%%%%%%%%%%%%%%%%
\begin{PromptIO}[GenRM Early Failed Example 2]{lst:qualitative_genrm_early_2}
[Question]
How do animals perceive the world, and what is the key to understanding their behavior?

[Parsed Final Answer]
\$ANSWER

[Reference Answer]
Animals perceive the world through details, and understanding this detail-oriented perception is the key to their behavior.

[GenRM Verification]
### Step 1: Identify the key points in the reference answer  
- Animals perceive the world through **details**.  
- **Understanding this detail-oriented perception** is the **key to their behavior**.

### Step 2: Check if the candidate answer contains all these key points  
The candidate answer is:  
**\"Animals perceive the world through details, and understanding this detail-oriented perception is the key to their behavior.\"**
This answer **contains all the key points** from the reference answer.

### Step 3: Verify the factual accuracy of the candidate answer  
The candidate answer is **factually accurate** and **does not contain any errors**. It restates the reference answer in the same terms.

### Step 4: Assess whether the candidate answer adequately addresses the question  
The question is:  
**\"How do animals perceive the world, and what is the key to understanding their behavior?\"**

The candidate answer **adequately addresses the question** by explaining that animals perceive the world through details and that understanding this detail-oriented perception is the key to their behavior.

### Final Decision:  
Final Decision: \textbackslash boxed\{Yes\}
\end{PromptIO}

%%%%%%%%%
\begin{PromptIO}[GenRM Failed Example: False Negative]{lst:qualitative_genrm_FN}
[Question]
What is the medical definition of a leiomyoma?

[Parsed Final Answer]
a benign neoplasm composed of smooth muscle cells

[Reference Answer]
A **leiomyoma** is a **benign tumor composed of smooth muscle cells**, most commonly found in the uterus (as fibroids), skin, or gastrointestinal tract. It is non-cancerous, slow-growing, and may or may not cause symptoms depending on its location and size.

[GenRM Verification]
### Step 1: Identify the key points in the reference answer
- A **leiomyoma** is a **benign tumor**.
- It is **composed of smooth muscle cells**.
- It is **most commonly found in the uterus (as fibroids)**, **skin**, or **gastrointestinal tract**.
- It is **non-cancerous**.
- It is **slow-growing**.
- It **may or may not cause symptoms** depending on its location and size.

### Step 2: Check if the candidate answer contains all these key points
- The candidate answer states: **\"a benign neoplasm composed of smooth muscle cells\"**.
- It includes the key point that it is a **benign tumor** (or **neoplasm**).
- It includes the key point that it is **composed of smooth muscle cells**.
- It **does not mention** the common locations (uterus, skin, gastrointestinal tract).
- It **does not mention** that it is **non-cancerous**, **slow-growing**, or that it **may or may not cause symptoms**.

### Step 3: Verify the factual accuracy of the candidate answer
- The candidate answer is **factually accurate** in stating that a leiomyoma is a **benign neoplasm composed of smooth muscle cells**.
- However, it **omits important contextual information** that is present in the reference answer.

### Step 4: Assess whether the candidate answer adequately addresses the question
- The question asks for the **medical definition** of a leiomyoma.
- While the candidate answer provides a **correct** definition, it is **incomplete** compared to the reference answer.
- The reference answer provides a **more comprehensive** and **accurate** medical definition, including key contextual details.

### Final Decision:
The candidate answer is **factually accurate** but **incomplete** and **less comprehensive** than the reference answer. It **misses important key points** such as the common locations, non-cancerous nature, slow-growing nature, and potential for symptoms.

Final Decision: \textbackslash boxed\{No\}
\end{PromptIO}

%%%%%%%%%%%%%%
\begin{PromptIO}[GenRM Failed Example: False Positive]{lst:qualitative_genrm_FP}
[Question]
What is the clinical significance of a decreased hemoglobin (Hgb) level, and what conditions might this indicate?

[Parsed Final Answer]
"Conditions that might indicate a decreased hemoglobin level include iron deficiency anemia, vitamin B12 or folate deficiency and bone marrow disorders.

[Reference Answer]
A decreased hemoglobin (Hgb) level may indicate: Anemia, Blood loss, Chronic diseases, or Hemolysis.

[GenRM Verification]
1. **Completeness and Meaningfulness**: The candidate answer is complete and meaningful. It provides specific conditions that can lead to a decreased hemoglobin level, rather than placeholders or incomplete expressions.

2. **Key Points in the Reference Answer**: The reference answer identifies four key conditions: Anemia, Blood loss, Chronic diseases, or Hemolysis.

3. **Containment of Key Points**: The candidate answer includes several conditions that are related to the reference answer. It includes:
- Iron deficiency anemia (a type of anemia)
- Vitamin B12 or folate deficiency (types of anemia)
- Bone marrow disorders (which can be related to chronic diseases or anemia)
While the reference answer lists \"Chronic diseases\" as a general category, the candidate answer includes specific examples that fall under this category (e.g., bone marrow disorders, which can be related to chronic conditions).

4. **Factual Accuracy**: The candidate answer is factually accurate. The listed conditions are all valid causes of decreased hemoglobin levels.

5. **Addressing the Question**: The candidate answer adequately addresses the question by listing specific conditions that might indicate a decreased hemoglobin level, which is what the question asks.

Final Decision: \textbackslash boxed\{Yes\}
\end{PromptIO}

%%%%%%%%%%%% Prompts
\section{LLM Prompts}
\label{appn: prompts}
We include all prompts we use for LLMs in this section. This include: prompt for final answer verification, prompt for generating rubrics, prompt for evaluating one rubric item during RL.
\subsection{Prompt for Final Answer Verification}
\label{appn: prompt final answer}
We use the prompt in Listing~\ref{lst:appn_prompt_answer_verify} to verify final answer for open-ended questions. This is used when: inference question difficulty for Dr.\ SCI, verifying final answer in SciRubric-Guided RL, verifying final answer using GenRM in Section~\ref{sec: ablation Open-Ended} and evaluating models on GPQA-General. ``<QUESTION>", ``<REF\_ANSWER>" and ``<GEN\_ANSWER>" are placeholders for question, reference answer, and model generated final answer during verification.

\begin{lstlisting}[style=prompt,caption={Answer Verification Prompt},label={lst:appn_prompt_answer_verify}]
You are a strict academic evaluator. Compare the candidate answer with the reference answer to determine if they are equivalent in correctness and completeness.

First, analyze the answers step by step:
1. Check if the candidate answer is complete and meaningful (not just placeholders, variables, or incomplete expressions)
2. Identify the key points in the reference answer
3. Check if the candidate answer contains all these key points
4. Verify the factual accuracy of the candidate answer
5. Assess whether the candidate answer adequately addresses the question

The candidate answer should be considered correct ONLY if:
- It is a complete, meaningful answer (not just placeholders like "$ANSWER", "X", "$", or similar)
- It contains all the key points from the reference answer
- The information is factually accurate
- It adequately addresses the question asked

Answer "No" if the candidate answer:
- Is just a placeholder, variable, or incomplete expression (e.g., "$ANSWER", "X", "$", "ANSWER", etc.)
- Is missing important key points from the reference answer
- Contains factual errors or inaccuracies
- Is significantly incomplete compared to the reference
- Uses different terminology that changes the meaning
- Only partially addresses the question
- Is empty, contains only whitespace, or is clearly malformed

Be strict in your evaluation. When in doubt, answer "No". Pay special attention to placeholder-like answers that appear to be formatting artifacts rather than actual solutions.

After your analysis, provide your final decision in the format: Final Decision: \\boxed{{Yes}} or Final Decision: \\boxed{{No}}

## Question:
<QUESTION>

## Reference Answer: 
<REF_ANSWER>

## Candidate Answer: 
<GEN_ANSWER>
\end{lstlisting}

\subsection{Prompt for Rubric Generation}
\label{appn: prompt rubric gen}
We use the prompt in Listing~\ref{lst:appn_prompt_rubric_gen} to generate rubrics for each open-ended question during construction of Dr.\ SCI dataset. The prompt is borrowed from ~\citet{gunjal2025rubrics} with some modifications. ``<SUBJECT>", ``<QUESTION>" and ``<REF\_ANSWER>" are placeholders for the subject, question, and reference answer for every open-ended instance in Dr.\ SCI dataset.

\begin{lstlisting}[style=prompt,caption={Rubric Generation Prompt},label={lst:appn_prompt_rubric_gen}]
You are an expert rubric designer for scientific reasoning questions. Your job is to generate a self-contained set of evaluation criteria or "rubrics" for judging how good a response is to a given question in one of STEM subjects (math, physics, chemistry, biology, medicine, cs, economics). Rubrics should  cover aspects such as factual correctness, depth of reasoning, clarity, logic correctness, completeness, style, helpfulness, and common pitfalls. Each rubric item must be fully self-contained so that non-expert readers need not consult any external information.

\\textbf{{Inputs:}}
\\begin{{itemize}}
  \\item \\texttt{{subject}}: <SUBJECT>
  \\item \\texttt{{question}}: <QUESTION>
  \\item \\texttt{{reference_answer}}: {REF_ANSWER}
\\end{{itemize}}

\\textbf{{Total items:}}
\\begin{{itemize}}
  \\item Choose 7-10 rubric items based on question complexity.
\\end{{itemize}}

Each rubric item must include exactly three keys:
\\begin{{enumerate}}
  \\item \\textbf{{title}}: 2-4 words summarization
  \\item \\textbf{{description}}: One sentence explicitly stating what to look for. For example:
    \\begin{{itemize}}
      \\item States that in the described closed system, the total mechanical energy (kinetic plus potential) before the event equals the total mechanical energy after the event.
      \\item Breaks down numerical energy values for each stage, demonstrating that initial kinetic energy plus initial potential energy equals final kinetic energy plus final potential energy.
      \\item Provides a concrete example, such as a pendulum converting between kinetic and potential energy, to illustrate how energy shifts within the system.
      \\item Does not mention that frictional or air-resistance losses are assumed negligible when applying conservation of mechanical energy.
    \\end{{itemize}}
  \\item \\textbf{{category}}: one from "Essential", "Important", "Optional", or "Pitfall" indicating the type of the rubric item
\\end{{enumerate}}

\\textbf{{Category guidance:}}
\\begin{{itemize}}
  \\item Essential: critical fact or step; omission invalidates the
answer.
  \\item Important: key information or reasoning; absence
severely weakens the response.
  \\item Optional: secondary details or actions; doesn't directly
affects correctness.
  \\item Pitfall: common but vital mistakes; must be penalized
if exist.
\\end{{itemize}}

\\textbf{{Format notes:}}
\\begin{{itemize}}
  \\item When referring to answer choices, explicitly say "Identifies (A)", "Identifies (B)", etc.
  \\item If a clear conclusion is required (e.g. "The final answer is (B)"), include an Essential Criteria for it.
  \\item If reasoning should precede the final answer, include an Important Criteria to that effect.
  \\item If brevity is valued, include an Optional Criteria about conciseness.
\\end{{itemize}}

\\textbf{{Output:}}  
Provide a JSON array of rubric objects as your final result after reasoning. Each object must contain exactly three keys-title, description, and category. Do not copy large blocks of the question or reference_answer into the text. Each description must begin with its category prefix, and no extra keys are allowed.

Now, given the question and reference_answer, generate the rubric as described. The reference answer is an ideal response but not necessarily exhaustive; use it only as guidance. You may try to solve the problem if you think it is necessary.'''
\end{lstlisting}

\subsection{Prompt for Evaluating 1 Rubric Item}
\label{appn: prompt rubric eval}
We use Listing~\ref{lst:appn_prompt_rubric_sys} as system prompt and Listing~\ref{lst:appn_prompt_rubric_user} as query template for reward model (Qwen3-4B) in our experiments to assess a response against one rubric item. ``<QUESTION>", ``<RUBRIC\_ITEM>" and ``<RESPONSE>" are placeholders for the question, a rubric item of this question, and a final response parsed from a model's rollout for this questions as introduced in Section~\ref{sec: method rubric}.

\begin{lstlisting}[style=prompt,caption={System Prompt for Evaluating a Rubric Item},label={lst:appn_prompt_rubric_sys}]
You are an academic evaluator verifying whether a candidate response meets a specific rubric criterion.

**Task**: Provide a binary verification (Yes/No) on whether the response satisfies the given rubric item.

**Rubric Criterion Types**:
1. **Essential**: Critical requirements that must be present for a good response
2. **Important**: Significant requirements that should be present for quality
3. **Optional**: Nice-to-have requirements that enhance response quality
4. **Pitfall**: Common mistakes or faults that should NOT be present in the response

**Evaluation Instructions by Type**:
- **For Essential/Important/Optional criteria**: Check if the response demonstrates the required positive behavior or includes the specified element. Output "Yes" if the good behavior is present, "No" if absent.
- **For Pitfall criteria**: Check if the response contains the specified fault or bad behavior. Output "Yes" if the fault EXISTS (response fails this criterion), "No" if the fault does NOT exist (response passes this criterion).

**Evaluation Guidelines**:
1. **Focus**: Only evaluate the specific rubric criterion - not overall correctness or other aspects
2. **Evidence Required**: Look for explicit evidence that demonstrates compliance with the rubric requirement
3. **Standards**: The response must explicitly demonstrate the required element (for positive criteria) or explicitly avoid the specified fault (for pitfall criteria)
4. **Quality Indicators**: Clear reasoning, proper application of specified approaches, conscious addressing of the criterion

**Response Format**:
- Brief analysis of how the response meets (or fails to meet) the rubric criterion
- For Pitfall criteria: Clearly state whether the specified fault is present or absent
- Focus only on the specified rubric item
- Conclude with: Final Decision: \\boxed{{Yes}} or Final Decision: \\boxed{{No}}
\end{lstlisting}

\begin{lstlisting}[style=prompt,caption={Query Template for Evaluating a Rubric Item},label={lst:appn_prompt_rubric_user}]
Given the following question, rubric criterion, and candidate response, please rate whether the response satisfies the rubric criterion with a binary decision (Yes/No).

# Question:
<QUESTION>


# Rubric:
<RUBRIC_ITEM>


# Response:
<RESPONSE>

\end{lstlisting}

%%%%%%%%%%%%%%%%%%%%%%%%%%%%%%%%%%%%%%%%%%%%%%%%%%%%%%%%%%%%%%%%%%%%%%%%%%%%%%%
%%%%%%%%%%%%%%%%%%%%%%%%%%%%%%%%%%%%%%%%%%%%%%%%%%%%%%%%%%%%%%%%%%%%%%%%%%%%%%%

\end{document}